\documentclass[11pt]{article}
\usepackage{amsmath,amsfonts}
\usepackage{graphicx}
\usepackage{lscape}
\usepackage{multirow}
\usepackage[sort]{natbib}
\bibpunct{(}{)}{;}{a}{,}{,}

\def\Var{\text{Var}}
\def\Cov{\text{Cov}}
\def\MSE{\text{MSE}}
\def\NLPD{\text{NLPD}}

\begin{document}

\begin{center}

\vspace*{0.8in}

\Large\textbf{Gaussian Process Regression with \\[3pt]
              Heteroscedastic or Non-Gaussian Residuals} \\[20pt]

\begin{minipage}{0.45\linewidth}
\centering
\small
Chunyi Wang\\
Department of Statistical Sciences\\
University of Toronto\\
\texttt{chunyi@utstat.toronto.edu}\\

\end{minipage}
\begin{minipage}{0.45\linewidth}
\centering
\small
Radford M.\ Neal\\
Department of Statistical Sciences and \\
Department of Computer Science\\
University of Toronto\\
\texttt{radford@utstat.toronto.edu}
\end{minipage}
\\
\vspace{0.2in}
\normalsize
26 December 2012
\end{center}

\vspace{0.1in}

\noindent \textbf{Abstract} Gaussian Process (GP) regression models
typically assume that residuals are Gaussian and have the same
variance for all observations. However, applications with
input-dependent noise (heteroscedastic residuals) frequently arise in
practice, as do applications in which the residuals do not have a
Gaussian distribution. In this paper, we propose a GP Regression model
with a latent variable that serves as an additional unobserved
covariate for the regression.  This model (which we call GPLC) allows
for heteroscedasticity since it allows the function to have a changing
partial derivative with respect to this unobserved covariate. With a
suitable covariance function, our GPLC model can handle (a) Gaussian
residuals with input-dependent variance, or (b) non-Gaussian residuals
with input-dependent variance, or (c) Gaussian residuals with constant
variance. We compare our model, using synthetic datasets, with a model
proposed by \citet*{Goldberg:1998}, which we refer to as GPLV, which
only deals with case (a), as well as a standard GP model which can
handle only case (c). Markov Chain Monte Carlo methods are developed
for both modelsl.  Experiments show that when the data is
heteroscedastic, both GPLC and GPLV give better results (smaller mean
squared error and negative log-probability density) than standard GP
regression. In addition, when the residual are Gaussian, our GPLC
model is generally nearly as good as GPLV, while when the residuals
are non-Gaussian, our GPLC model is better than GPLV.

\begin{section}{Introduction}

Gaussian Process (GP) regression models are popular in the machine
learning community (see, for example, the text by \cite{gpml}),
perhaps mainly because these models are very flexible --- one can
choose from many covariance functions to achieve different degrees of
smoothness or different degrees of additive structure, the parameters
of such a covariance function can be automatically determined from the
data. However, standard GP regression models typically assume that the
residuals have i.i.d.\ Gaussian distributions that do not depend on
the input covariates, though in many applications, the variances of
the residuals do depend on the inputs, and the distributions of the
residuals are not necessarily Gaussian. In this paper, we present a GP
regression model which can deal with input-dependent residuals. This
model includes a latent variable with a fixed distribution as an
unobserved input covariate.  When the partial derivative of the
response with respect to this unobserved covariate changes across
observations, the variance of the residuals will change. When the
latent variable is transformed non-linearly, the residuals will be
non-Gaussian.  We call this the ``Gaussian Process with a Latent
Covariate'' (GPLC) regression model.


In Section \ref{sec:models} below, we give the details of this model
as well as the standard GP model (``STD'') and a model due to
\citet*{Goldberg:1998}, which we call the ``Gaussian Process
regression with a Latent Variance'' (GPLV) model, and we discuss the
relationships/equivalencies between these models. We describe
computational methods in Section \ref{sec:computation}, and present
the results of these models on various synthetic datasets in Section
\ref{sec:exp}.  

\end{section}

\begin{section}{The models}\label{sec:models}

We look at non-linear regression problems, where the aim is to find the 
association between a vector of covariates $x$ and a response $y$ using $n$ 
observed pairs $(x_1,y_1),...,(x_n,y_n)$, and then make predictions for
$y_{n+1},y_{n+2},...$ corresponding to $x_{n+1},x_{n+2}...$:
	\begin{equation}
		\label{reg}
		y_i = f(x_i) + \epsilon_i
	\end{equation}
The covariate $x_i$ is a vector of length $p$, and the
correspoding response $y_i$ is a scalar.

\begin{subsection}{The standard GP regression model}

In the standard Gaussian process regression model The random
residuals, $\epsilon_i$'s, are assumed to have i.i.d.\ Gaussian
distributions with mean 0 and constant variance $\sigma^2$.
 
Bayesian GP models assume that the noise-free function $f$ comes from
a Gaussian Process which has prior mean function zero and some
specified covariance function. Note that a zero mean prior is not a
requirement --- we could specify a non-zero prior mean function $m(x)$
if we have \textit{a priori} knowledge of the mean
structure. Using a zero mean prior just reflects our prior knowledge
that the function is equally likely to be positive or
negative. It doesn't mean we believe the actual function will have 
an average over its domain of zero.

The covariance functions can be fully-specified functions, but common
practice is to specify a covariance function with
unknown hyperparameters, and then estimate the hyperparameters from
the data.  Given the values of the hyperparameters, the 
responses, $y$, in a set of cases have a multivariate Gaussian distribution 
with zero mean and a covariance matrix given by
	\begin{equation}
		\label{covy}
		\Cov(y_i,y_j) = K(x_i,x_j)  + \delta_{ij}\sigma^2
	\end{equation}
where $\delta_{ij}=0$ when $i\ne j$ and $\delta_{ii}$ = 1, and
$K(x_i,x_j)$ is the covariance function of $f$. Any function that
always leads to a positive semi-definite covariance matrix can be used as a
covariance function.  One example is the squared exponential
covariance function with isotropic length-scale:
	\begin{equation}
		\label{eq:covSEiso}
		K(x_i,x_j) = c^2 +  \eta^2 \exp\left( -\frac{\|x_{i}-x_{j}\|^2}{\rho^2}\right)
	\end{equation}
Where $c$ is a suitably chosen constant, and $\eta, \rho$ and $\sigma$ are
hyperparameters. $\eta^2$ is sometimes referred to as the ``signal
variance'', which controls the magnitude of variation of $f$; $\sigma^2$ is the
residual variance; $\rho$ is the length scale parameter for the input
covariates. We can also assign a different length scale parameter to
each covariate, which leads to the squared exponential covariance
function with automatic relevance determination (ARD):
	\begin{equation}
		\label{covfun}
		K(x_i,x_j) = c^2 +  \eta^2 \exp\left( -\sum_{k=1}^p \frac{(x_{ik}-x_{jk})^2}{\rho_k^2}\right)
	\end{equation}
We will use squared exponential forms of covariance function from
\eqref{eq:covSEiso} or \eqref{covfun} in most of this paper.

If the values of the hyperparameters are known, then the predictive
distribution of $y_*$ for a test case $x_*$ based on observed
values $x=(x_1,...,x_n)$ and $y=(y_1,...,y_n)$ is Gaussian with the
following mean and variance:
	\begin{equation}
		\label{predmean}
		E(y_*|x,y,x_*) = k^TC^{-1}y
	\end{equation}
	\begin{equation}
		\label{predvar}
		\Var(y_*|x,y,x_*) = v - k^TC^{-1}k
	\end{equation}
In the equations above, $k$ is the vector of covariances between
$y_*$ and each of $y_1,\ldots,y_n$, $C$ is the covariance matrix of the
observed $y$, and $v$ is the prior variance of $y_*$, which is
$\Cov(y_*,y_*$) from \eqref{covy}. 


When the values of the hyperparameters (denoted as $\theta$) are
unknown and therefore have to be estimated from the data, we put
priors on them (typically independent Gaussian priors on the logarithm
of each hyperparameter), and obtain the posterior distribution
$p(\theta|x,y) \propto \mathcal{N}(y|0,C(\theta))p(\theta)$. The predictive
mean of $y$ can then be computed by integrating over the posterior
distribution of the hyperparameters:
	\begin{equation}
		\label{predmeanint}
		E(y_*|x,y,x_*) = \int_{\Theta} k(\theta)^TC(\theta)^{-1}y \cdot p(\theta|x,y) d\theta
	\end{equation}
Letting $\mathcal{E} = E(y_*|x,y,x_*)$, the predictive variance is given by
	\begin{align}
		\label{predvarint}
		\Var(y_*|x,y,x_*) &= E_{\theta}[\Var(y_*|x,y,x_*,\theta)] +\Var_{\theta}[E(y_*|x,y,x_*,\theta)] \\[3pt]\notag
		&=\int_\Theta \left[v(\theta) - k(\theta)^TC(\theta)^{-1}k(\theta)\right]p(\theta|x,y)d\theta + 
			\int_\Theta \left[ k(\theta)^TC(\theta)^{-1}y - \mathcal{E} \right]^2 p(\theta|x,y) d\theta
	\end{align}



\end{subsection}

\begin{subsection}{A GP regression model with a latent covariate }

In this paper, we consider adding a latent variable $w$ into the model
as an unobserved input. The regression equation then becomes
	\begin{equation}
		\label{lat_reg}
		y_i= g(x_i,w_i) + \zeta_i.
	\end{equation}
In this setting, the latent value $w_i$ has some known random
distribution, the same for all $i$.  If we view $g(x_i,w_i)$ 
as a function of $x_i$ only, its value is random, due to the
randomness of $w_i$.  So $g$ is not the regression function 
giving the expected value of $y$ for a given value of $x$ --- 
that is given by the averge value of $g$ over all $w$, which
we write as $f(x) = E(y|x)$:
	\begin{equation}
		f(x) = \int g(x,w)p(w) dw
		\label{eq:avgfun}
	\end{equation}
where $p(w)$ is the probability density of $w$. Note that
\eqref{eq:avgfun} implies that the term $\zeta_i$, which we assume has
i.i.d.\ Gaussian distribution with constant variance, is not the real
residual of the regression, since
	\[
		\zeta_i = y_i - g(x_i,w_i)\ \ne\ y_i - f(x_i) = \epsilon_i
	\]
where $\epsilon_i$ is the true residual. We put $\zeta_i$ in the
regression for two reasons.  First, the covariance function for $g$
can sometimes produce nearly singular covariance matrices, that are
computationally non-invertible because of round-off error. Adding a small 
diagonal term can avoid the computational issue without significantly changing
the properties of the covariance matrix. Secondly, the function $g$
will produce a probability density function for $\epsilon$ that has
singularities at points where the derivative of $g$ with respect to
$w$ is zero, which is probably not desired in most applications.
Adding a jitter term $\zeta_i$ smooths away such singularities.

We will model $g(x,w)$ using a GP with a squared exponential covariance 
function with ARD, for which
the covariance between training cases $i$ and $j$, with latent
values $w_i$ and $w_j$, is
	\begin{align}
		\label{eq:cov_gplc}
		\Cov(y_i,y_j) &= K((x_i,w_i),(x_j,w_j)) + \sigma^2\delta_{ij} 
                \\[3pt] \notag 
		&=c^2 + \eta^2 \exp\left(-\sum_{k=1}^p \frac{(x_{ik}-x_{jk})^2}{\rho_k^2} 
			-  \frac{(w_i-w_j)^2}{\rho_{p+1}^2} \right) + \sigma^2\delta_{ij}
	\end{align}

We choose independent standard normals as the distributions for
$w_1,...,w_n$.  The mean for the $w_i$ is chosen to be zero because
the squared exponential function is stationary, and hence only the
difference between $w_i$ and $w_j$ matters. The variance of the $w_i$
is fixed at 1 because the effect of a change of scale of $w_i$ can be
achieved instead by a change in the length scale parameter $l_{p+1}$.

We write $p(w)$ for the density for the vector of latent variables $w$, and
$p(\theta)$ for the prior density of all the hyperparameters (denoted as a
vector $\theta$). The posterior joint density for the latent variables
and the hyperparameters is
	\begin{equation}
		p(w,\theta|x,y) = \mathcal{N}(y|0,C(\theta, w)) p(w)p(\theta)
	\end{equation}
where $\mathcal{N}(a|\mu,\Sigma)$ denotes the probability density of a
multivariate Gaussian distribution with mean $\mu$ and covariance
matrix $\Sigma$, evaluated at $a$. $C(\theta, w)$ is the covariance
matrix of $y$, which depends on $\theta$ and $w$.  

	
The prediction formulas for GPLC models are similar to
\eqref{predmeanint} and \eqref{predvarint}. In addition to averaging
over the hyperparameters, we also have to average over the posterior
distribution of the latent variables $w=(w_1,...,w_n)$:
	\begin{equation}\label{eq:predmeangplc}
		E(y_*|x,y,x_*) = \int_{\mathcal{W}}\int_{\Theta} k(\theta,w,w_*)^TC(\theta,w)^{-1}y\, p(\theta,w|x,y) d\theta dw
	\end{equation}
	\begin{align}\label{eq:predvargplc}
		\Var(y_*|x,y,x_*) &= 
		E_{\theta,w}[\Var(y_*|x,y,x_*,\theta,w)] +\Var_{\theta,w}[E(y_*|x,y,x_*,\theta,w)] \\[3pt]\notag
		&=\int_{\mathcal{W}}\int_\Theta \left[ v(\theta,w_*) - k(\theta,w,w_*)^T C(\theta,w)^{-1} k(\theta,w,w_*) \right] p(w,\theta|x,y) d\theta dwdw_*\\[3pt]\notag
		& \ \ \ \ \ + \int_{\mathcal{W}}\int_\Theta \left[ k(\theta,w,w_*)^TC(\theta,w)^{-1}y - \mathcal{E} \right]^2 p(w,\theta|x,y)d\theta dwdw_*
	\end{align}
where $\mathcal{E} = E(y_*|x,y,x_*)$

Note that the vector of covariances of the response in a test case
with the responses in training cases, written as $k(\theta,w)$ in
\eqref{eq:predmeangplc} and \eqref{eq:predvargplc}, depends on, $w_*$,
the latent value for the test case. Since we do not observe $w_*$, we
randomly draw values from the prior distribution of $w_*$, compute the
corresponding expectation or variance and take the average.  Similarly,
the prior variance for the response in a test case, written $v(\theta,w_*)$
above, depends in general on $w_*$ (though not for the squared
exponential covariance function that we use in this paper).
	
To see that this model allows residual variances to depend on
$x$, and that the residuals can have non-Gaussian distributions, we
compute the Taylor-expansion of $g(x,w)$ at $w=0$:
	\begin{equation}
		\label{f_taylor}
		g(x,w) = g(x,0) + g_2'(x,0)w + \frac{w^2}{2}g_2''(x,0) + ...
	\end{equation}
where $g_2'$ and $g_2''$ denotes the first and second order partial
derivatives of $g$ with respect to its second argument ($w$). If we can
ignore the second and higher order terms, i.e. the linear
approximation is good enough, then the response given $x$ is Gaussian,
and
	\begin{align}
		\label{f_approx_var}
		 \Var[g(x,w)] &\approx 0 + [g_2'(x,0)]^2 \Var(w) = [g_2'(x,0)]^2
	\end{align}
which depends on $x$ when $g_2'(x,0)$ depends on $x$ (which
usually is the case when $g$ is drawn from a GP prior). Thus in this case, 
the model produces Gaussian residuals with input-dependent variances.

\begin{figure}[t]
	\centering
	\includegraphics[width=3.5in, height=2.8in]{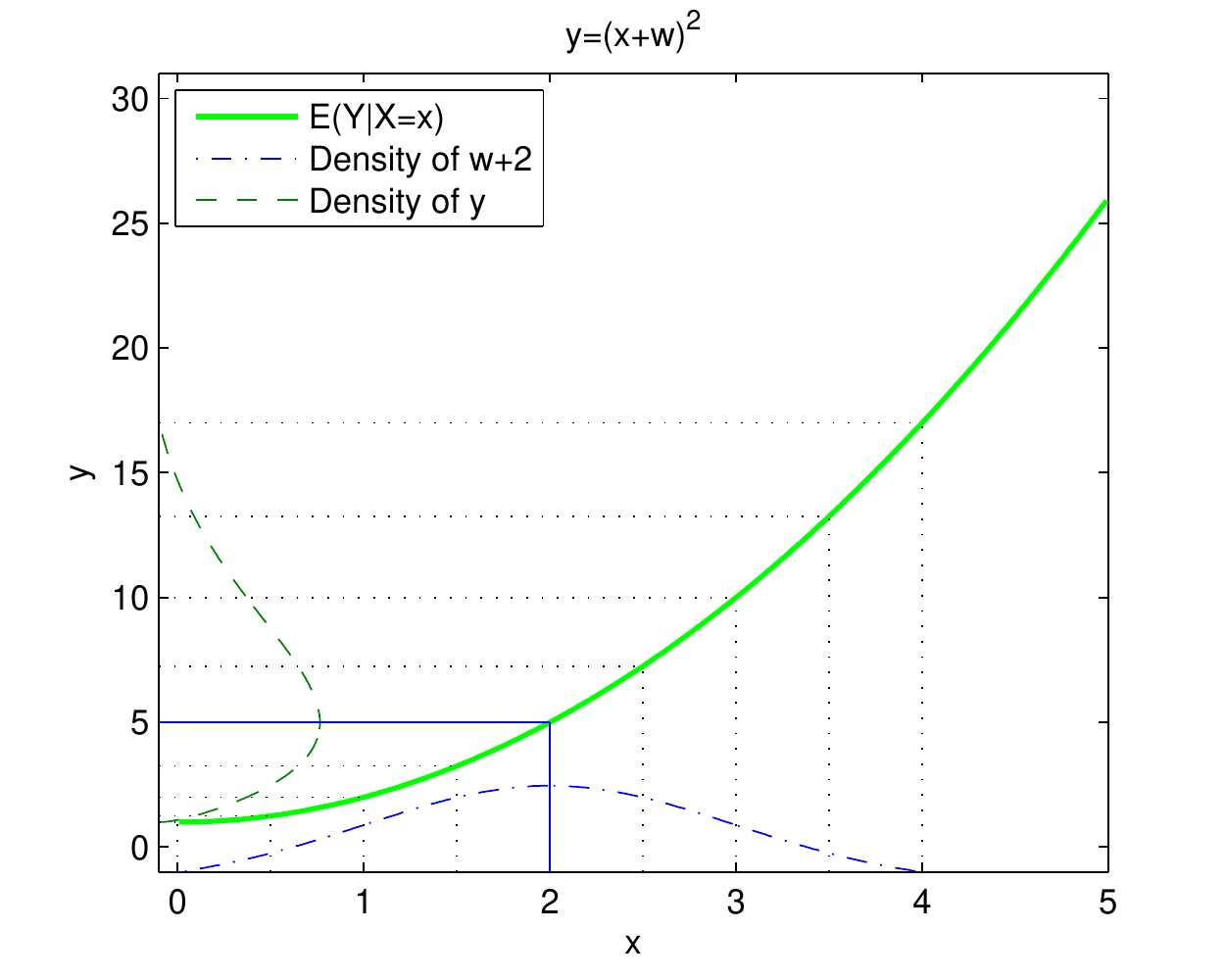}
	\caption{How GPLC can produce a non-Gaussian distribution of residuals.}
	\label{fig:trans}
\end{figure}	

If the high-order terms in \eqref{f_taylor} cannot be ignored, then
the model will have non-Gaussian, input-dependent residuals. For
example, consider $g(x,w)=(x+w)^2$, where the second order term in $w$
clearly cannot be ignored. Conditional on $x$, $g (x,w)$ follows a
non-central Chi-Squared distribution. Figure \ref{fig:trans}
illustrates that at $x=2$, an unobserved normally distributed input
$w$ translates into a non-Gaussian output $y$. Note that for
demonstration purposes the density curves of $w$ and $y$ are not to
scale (since the scales on the $x$-axis and $y$-axis are different).

	\begin{figure}[t]
		\centering
		\begin{minipage}{0.45\linewidth}
			\flushright
			\includegraphics[width=3in,height=3in]{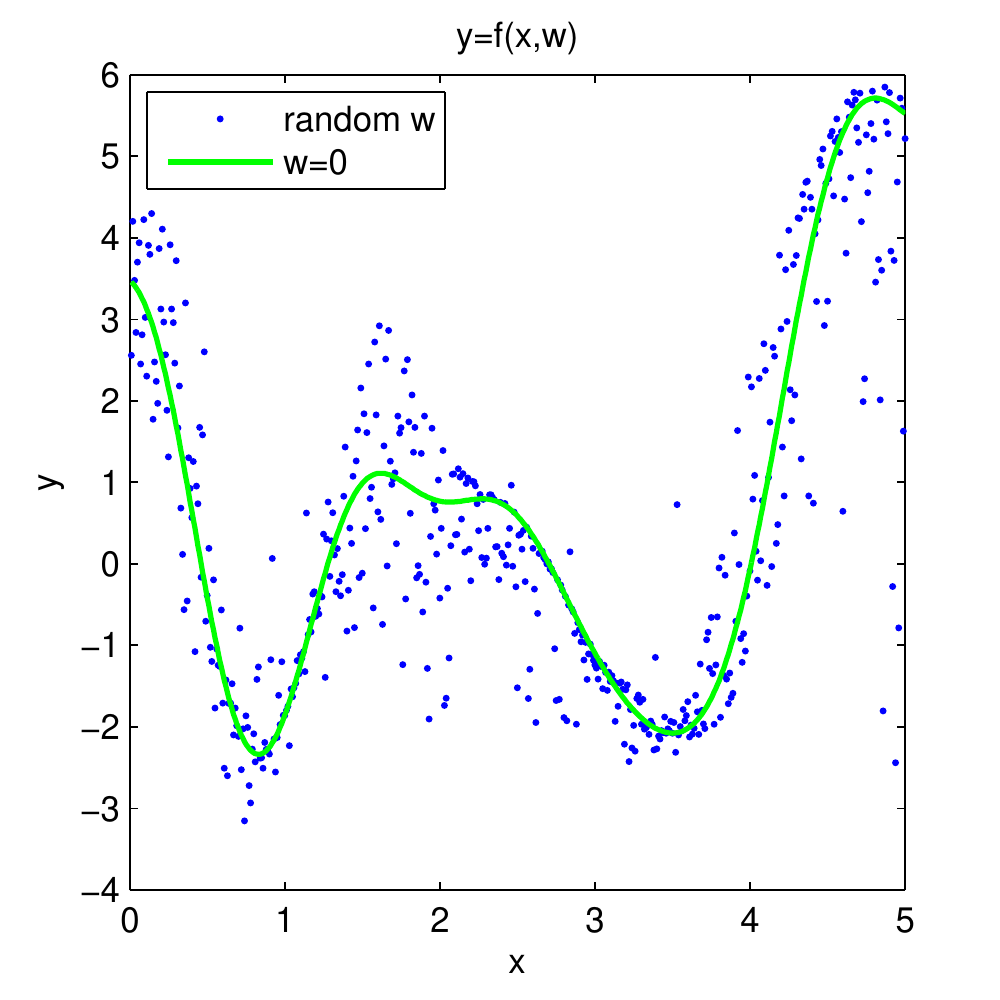}
		\end{minipage}
		\hspace{0.1in}
		\begin{minipage}{0.45\linewidth}
			\includegraphics[width=3in,height=3in]{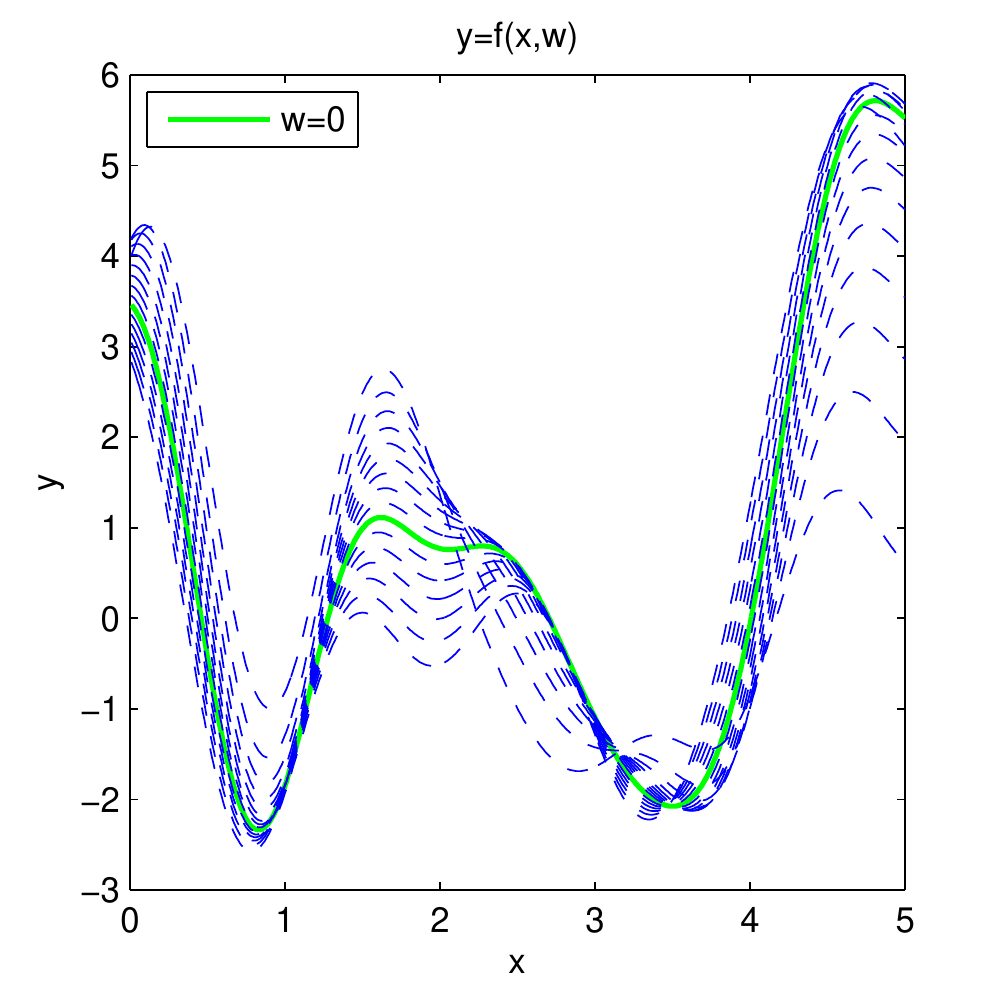}
		\end{minipage}	
		\caption{Heteroscedasticity produced by an unobserved covariate.
The left plot shows a sample of $x$ and $y$ from the GP prior, with $w$ not
shown.  The right plot shows 19 dashed curves of $g(x_i,w_i)$ (for the same
$g$ as on the left) where the $w_i$ are fixed to the same value, equal
to the $5i$th percentile of the standard normal for the $i$th curve
(i.e.\ 1 to 19).}
		\label{fig:lateff}
	\end{figure}

Figure \ref{fig:lateff} illustrates how an unobserved covariate can
produce heteroscedasticity. The data in the left plot are generated
from a GP, with $x_i$ drawn uniformly from [0,5] and $w_i$ drawn from
$N(0,1)$.  The hyperparameters of the squared exponential covariance
function were set to $\eta=3$, $\rho_x=0.8$, and $\rho_w=3$. Supposing
we only observe $(x,y)$, the data is clearly heteroscedastistic, since
the spread of $y$ against $x$ changes when $x$ changes. For instance,
the spread of $y$ looks much bigger when $x$ is around 1.8 than it is
when $x$ is near 3.5. We also notice that the distribution of the
residuals can't be Gaussian, as, for instance, we see strong skewness
near $x=5$. These plots show that if an important input quantity is
not observed, the function values based only on the observed inputs
will in general be heteroscedastic, and non-Gaussian (even if the
noise term $\zeta_i$ is Gaussian).

Note that although an unobserved input quantity will create
heteroscedasticity, our model can work well even if no such quantity
really exists. The model can be seen as just using the latent
variable as a mathematical trick, to produce changing
residual variances. Whether or not there really exists an unobserved
input quantity doesn't matter (though in practice, unobserved
quantities often do exist).

\end{subsection}
\begin{subsection}{A GP regression model with a latent variance }

\citet*{Goldberg:1998} proposed a GP treatment of regression with
input-dependent residuals. In their scheme, a ``main'' GP models the
mean of the response just like a standard GP regression model, except
the residuals are not assumed to have constant variance --- a
secondary GP is used to model the logarithm of the standard deviation
of the noise, which depends on the input. The regression equation
looks the same as in \eqref{reg}:
	\begin{equation}
		\label{gb_reg}
		y_i = f(x_i) + \epsilon_i
	\end{equation}
but the residuals $\epsilon_1,...\epsilon_n$ are do not have the
same variance --- instead,
the logarithm of the standard deviation $z_i=\log SD[\epsilon(x_i)]$
depends on $x_i$ through:
	\begin{equation}
		z_i = r(x_i) + J_i
		\label{gb_sd}
	\end{equation}
$f(x)$ and $r(x)$ are both given (independent) GP priors, with zero
mean and covariance functions $C_y$ and $C_z$, which have different
hyperparameters (e.g.\ $(\eta_y, \rho_{y})$ and $(\eta_z, \rho_{z})$).
 $J_i$ is a Gaussian ``jitter'' term \citep[see][]{Neal:1997} which has
i.i.d.\ Gaussian distribution with zero mean and standard deviation $\sigma_J$ 
(a preset constant, usually a very small number, e.g.\ $10^{-3}$). Writing
$x=(x_1,...,x_n)$, $y=(y_1,...,y_n)$, $\theta_y = (\eta_y, \rho_{y})$,
$\theta_z=(\eta_z, \rho_{z})$, and $z=(z_1,...,z_n)$, the posterior
density function of the latent values and the hyperparameters is
	\begin{equation}
	\label{gplv_posterior}
	p(\theta_y,\theta_z,z|x,y)
	\ \propto\ p(y|x,z,\theta_y)p(z|x,\theta_z)p(\theta_y,\theta_z)
	\ \propto\ \mathcal{N}(y|0,C_y(\theta_y,z))\mathcal{N}(z|0,C_z(\theta_z))p(\theta_y,\theta_z)\ \
	\end{equation}
where $C_y$ is the covariance matrix for $y$ (for the ``main'' GP),
$C_z$ is the covariance matrix for $z$ (for the ``secondary'' GP) and
$p(\theta_y,\theta_z)$ represents the prior density for the
hyperparameters (typically independent Gaussian priors for their logarithms).
The predictive mean can be computed in a similar
fashion as the prediction of GPLC, but instead of averaging over $w$, we
average over~$z$.  To compute the covariance vector $k$, we
need the value of $z_{n+1}$, which we can sample from
$p(z_{n+1}|z_1,...,z_n)$.

Alternatively, instead of using a GP to model the logarithm of the
residuals standard deviations, we can set the standard
deviations to the absolute values of a function modeled by a GP.  That
is, we let $SD(\epsilon_i) = |z_i|$, with $z_i = r(x_i)$. So the 
regression model can be written as
	\begin{equation}
		\label{gplvalt}
		y_i = f(x_i) + r(x_i)u_i
	\end{equation}
where $u_i \stackrel{\text{iid}}{\sim} N(0,1)$.

This is similar to modeling the log of the standard deviation with a
GP, but it does allow the standard devaition, $|z_i|$, to be zero,
whereas $\exp(z_i)$ is always positive, and it is less likely to
produce extremely large values for the standard deviation of a
residual.  A more general approach is taken by \citet{Wilson:2010},
who use a parameterized function to map values modeled by a GP to
residual variances, estimating the parameters from the data.

In the original paper by \citeauthor{Goldberg:1998}, a toy example was
given where the hyperparameters are all fixed, with only the latent
values sampled using MCMC. In this paper, we will take a full Bayesian
approach, where both the hyperparameters and the $z$ values are
sampled from \eqref{gplv_posterior}. In addition, we will discuss
fast computation methods for this model in Section
\ref{sec:computation}.  

\end{subsection}

\begin{subsection}{Relationships between GPLC and other models}

We will show in this section that GPLC can be equivalent to the a
standard GP regression model or a GPLV model, when the covariance
function is suitably specified.

Suppose the function $g(x,w)$ in \eqref{lat_reg} has the form 
		\begin{equation}
			g(x_i,w_i) = h(x_i) + \sigma w_i
			\label{gplc->reg}
		\end{equation}
where $w_i\sim N(0,1)$. If we only observe $x_i$ but not $w_i$, then
\eqref{gplc->reg} is a regression problem with unknown i.i.d.\
Gaussian residuals with mean 0 and variance $\sigma^2$, which is
equivalent to the standard GP regression model \eqref{reg}, if we give
a GP prior to $h$.  If we specify a covariance function that produces
such a $g(x,w)$, then if we set $\zeta_i=0$ (or equivalently, the
hyperparameter $\sigma^2=\Var(\zeta_i)=0$), our GPLC model will be
equivalent to the standard GP model. Below, we compute the covariance
between training cases $i$ and $j$ (with latent values $w_i$ and
$w_j$) to find out the form of the appropriate covariance function.

Let's put a GP prior with zero mean and covariance function
$K_1(x_i,x_j)$ on $h(x)$. As usual, the $w_i$ have independent $N(0,1)$
priors.  Since the values of $g(x,w)$ are a linear combination of independent
Gaussians, they will have a Gaussian process distribution, conditional on the 
hyperparameters.  Now the covariance between cases $i$
and $j$ is
		\begin{align}
			\notag \Cov[g(x_i,w_i),g(x_j,w_j)] &= E[(h(x_i)+\sigma w_i)(h(x_j)+\sigma w_j)] \\
			\notag&=E[h(x_i)h(x_j)] + \sigma^2w_iw_j\\
			&=K_1(x_i,x_j) + \sigma^2w_iw_j
		\end{align}
Therefore, if we put a GP prior on $g(x,w)$ with zero mean and covariance 
function 
		\begin{equation}\label{gplc->reggp}
			K[(x_i,w_i),(x_j,w_j)] = K_1(x_i,x_j) + \sigma^2w_iw_j
		\end{equation}
the results given by GPLC will be equivalent to standard GP
regression with covariance function $K_1$. In practice, if we are
willing to make the assumption that the residuals have equal variances
(or know this as a fact), this modified GPLC model is not useful, since
the complexity of handling latent variables computationally is unnecessary. 
However, consider a more general covariance function
\begin{equation}\label{gplc-se-lin}
	K[(x_i,w_i),(x_j,w_j)] = K_1[(x_i,w_i),(x_j,w_j)] + K_2[(x_i,w_i),(x_j,w_j)]
\end{equation}
where $K_1[(x_i,w_i),(x_j,w_j)] = \exp\left( - \sum_{k=1}^p
(x_{ik}-x_{jk})^2/\rho_k^2 - (w_i-w_j)^2/\rho_{p+1}^2 \right)$ is a
squared exponential covariance function with ARD, and
$K_2[(x_i,w_i),(x_j,w_j)] = \sum_{k=1}^p \gamma_k^2 x_{ik}x_{jk} +
\gamma_{p+1}^2 w_iw_j$ is a linear covariance function with ARD. Then
the covariance function \eqref{gplc->reggp} can be obtained as a
limiting case of \eqref{gplc-se-lin}, when $\rho_{p+1}$ goes to
infinity in $K_1$ and $\gamma_1,...,\gamma_p$ all go to
zero. Therefore, we could use this more general model, and
let the data choose whether to (nearly) fit the simpler 
standard GP model.


Similarly, if we believe that the function $g(x,w)$ is of the form
		\begin{equation}
			g(x,w) = h_1(x) + wh_2(x)
			\label{gplc->gplv model}
		\end{equation}
with $h_1$ and $h_2$ independently having Gaussian Process priors with 
covariance functions $K_1$ and $K_2$,
we can use a GPLC model with a covariance function of the form
		\begin{equation}
			K[(x_i,w_i),(x_j,w_j)]  = K_1(x_i,x_j) + w_iw_j K_2(x_i,x_j)
			\label{gplc->gplv}
		\end{equation}

Now consider the alternative GPLV model \eqref{gplvalt}: if we put
independent GP priors on $f(x_i)$ and $r(x_i)$, each with zero mean,
and covariance functions $K_1$ and $K_2$, respectively, then model
\eqref{gplvalt} is equivalent to the modified GPLC model above
with covariance function \eqref{gplc->gplv}.  The
hyperparameters of $K_1$ of both models should have the same posterior
distribution, as would the hyperparameter of $K_2$. Notice that the
two models have different latent variables: the latent variable in
GPLC, $w_i$, is the value of the $i$th (normalized) residual; the
latent variable in GPLV is $z_i=r(x_i)$, which is plus or minus the
standard deviation of the $i$th residual. 


\end{subsection}
\end{section}

\begin{section}{Computation} \label{sec:computation}

Bayesian inference for GP models is based on the posterior
distribution of the hyperparameters and the latent
variables. Unfortunately this distribution is seldom analytically
tractable. We usually use Markov Chain Monte Carlo to sample the
hyperparameters and the latent values from their posterior
distribution.

\begin{subsection}{Overview of methods}
	
Common choices of MCMC method include the classic Metropolis algorithm
\citep{Metropolis:1953} and slice sampling \citep{Neal:2003}. The
Metropolis sampler is easy to implement, but for high-dimensional
distributions, it is generally hard to tune. We can update all the
parameters at each iteration using a multivariate proposal
distribution (e.g.\ $N(0, D)$ where $D$ is diagonal), or we can update
one parameter at a time based on a univariate proposal distribution.
Either way, to contruct an efficient MCMC method we have to assign an
appropriate value for the proposal standard deviation (a ``tuning
parameter'') for each hyperparameter or latent variable so that the
acceptance rate on each variable is neither too big nor too small
(generally, between 20\% and 80\% is acceptable, though the optimal
value is typically unknown). There is generally no good way to find
out what tuning parameter value is the best for each variable other
than trial and error. For high-dimensional problems, tuning the chain
is very difficult. Using squared exponential covariance function, our
model GPLC has $D=n+p+3$ variables (including hyperparameters and
latent variables), and the GPLV model has $D=n+2p+2$ variables.

Slice sampling, on the other hand, although slightly more difficult to
implement, is relatively easier to tune. It does also have tuning
parameters (one can control the step-out size and the number of
step-outs), but the performance of the chain is not very sensitive to
the tuning parameters. Figure 2.7 of \citet{Thompson:2011}
demonstrates that step-out sizes from 1 to 1000 all lead to similar
computation time, while a change in proposal standard deviation from 1
to 1000 for a Metropolis sampler can result in a MCMC which is 10 to
100 times slower.  In this paper, we use univariate step-out slice
sampling for regular GP regression models and GPLC models. For GPLV,
since the latent values are highly correlated, regular Metropolis and
slice samplers do not work well. We will give a modified Metropolis
sampler than works better than both of these simpler samplers.

\end{subsection}

\begin{subsection}{Major computations for GP models}

Evaluating the log of the posterior probability density of a GP model
is typically dominated by computating the covariance matrix, $C$, and
finding the Cholesky decomposition of $C$, with complexities $pn^2$
and $n^3$, respectively.

For both standard GP models and GPLV models, updates of most of the
hyperparameters require that the covariance matrix $C$ be recomputed,
and hence also the Cholesky decomposition (denoted as chol$(C)$).  For
GPLC, when the $i$th latent variable is updated, most of $C$ is
unchanged, except for the $i$th row and $i$th column. This change
requires a rank-$n$ update on the Cholesky decomposition, which is
almost as costly as as finding the Cholesky decomposition for a new
$C$.

Things are slightly more complicated for GPLV, since the model
consists of two GPs, with two covariance matrices. When one of the
hyperparameters for the main GP (denoted as $\theta_y$) is changed,
the covariance matrix for the main GP, $C_y$, is changed, and thus
chol($C_y$) has to be recomputed. However, $C_z$, the covariance
matrix for the secondary GP, remain unchanged. When one of $\theta_z$,
the hyperparameters of the secondary GP is changed, $C_y$ (and
chol$(C_y)$) remain unchanged, but $C_z$ and chol$(C_z)$ must be
recomputed. When one of the latent values, say the $i$th, is changed,
$C_z$ remains unchanged as it only depends on $x$ and $\theta_z$, but
the $i$th entry on the diagonal of $C_y$ is changed. This minor change
to $C_y$ requires only a rank-1 update \citep{Sherman:1950}, with
complexity $n^2$.

We list the major operations for the GP models discussed in this paper
in Table \ref{tbl:gp_operations}.
	
	\begin{table}[t]
		\centering
		\small
		\begin{tabular}{c|c|c|c}
			\hline
			\multicolumn{2}{c|}{} & one hyperparameter & one latent variable \\ \hline
			\multirow{3}{*}{STD} & Operation	& $C$, chol($C$)			& - \\\cline{2-4}
								& Complexity		& $pn^2$, $n^3$		& - \\\cline{2-4}
								& \# of such operations & $p+2$			& - \\\hline
			\multirow{3}{*}{GPLC} & Operation	& $C$, chol($C$)			&  $1/n$ of $C$, all of chol($C$) \\ \cline{2-4}
								& Complexity		& $pn^2$, $n^3$		& $pn$, $n^3$ \\\cline{2-4}
								& \# of such operations & $p+3$			& $n$ \\\hline
			\multirow{3}{*}{GPLV} & Operation	& $C_y$, chol($C_y$) or $C_z$, chol($C_z$) & rank-1 update $C_y$ \\\cline{2-4}
								& Complexity & $pn^2$, $n^3$ & $n^2$ \\\cline{2-4}
								& \# of such operations & $2p+2$			& $n$ \\\hline
		\end{tabular}
	\caption{Major operations needed when hyperparameters
                 and latent variables change in GP models.}
		\label{tbl:gp_operations}
	\end{table}

\end{subsection}

\begin{subsection}{A modified Metropolis sampler for GPLV}  \label{sec:modified_metro}

\citet{Neal:1998} describes a method for updating latent variables in
a GP model that uses a proposal distribution that takes into account
the correlation information.   This method proposes to change the current 
latent values, $z$, to a $z'$ obtained by
	\begin{equation}
		z' = (1-a^2)^{1/2}z + a L u
		\label{met:val}
	\end{equation} 
where $a$ is a small constant (a tuning parameter, typically slightly
greater than zero), $L$ is the lower triangular Cholesky decomposition
for $C_z$, the covariance matrix for the $N(0,C_z(\theta_z))$ prior for
$z$, and $u$ is a random vector of i.i.d.\ standard normal values.
The transition from $z$ to $z'$ is reversible, and leaves the prior
for $z$ invariant.  Because of this, the Metropolis-Hastings
acceptance probability for these proposals depends only on the ratio
of likelihoods for $z'$ and $z$.

We will use this method to develop a sampling strategy for
GPLV. Recall the unnormalized posterior distribution for the
hyperparameters and latent values is given by
	\begin{equation*}
		p(\theta_y,\theta_z,z|x,y)=\mathcal{N}(y|0,C_y(\theta_y,z))\mathcal{N}(z|0,C_z(\theta_z))p(\theta_y,\theta_z)
	\end{equation*}
To obtain new values $\theta_y',\theta_z'$ and $z'$
based on current values $\theta_y,\theta_z$ and $z$, we can do the
following:\vspace{-5pt}
\begin{enumerate}

\item For each of the hyperparameters in $\theta_y$ (i.e.\ those
associated with the ``main'' GP), do an update of this hyperparameter
(for instance a Metropolis or slice sampling update).  Notice that for
each of these updates we need to recompute chol($C_y$), but not
chol($C_z$), since $C_z$ does not depend on $\theta_y$.

\item For each of the hyperparameters in $\theta_z$ (i.e.\ those
for the ``secondary'' GP):
\begin{enumerate}

\item Do an update of this hyperparameter
(e.g.\ with Metropolis or slice sampling).  We need to recompute
chol($C_z$) for this, but not chol($C_y$), since $C_y$ does
not depend on $\theta_z$.

\item Update all of $z$ with the proposal described in
\eqref{met:val}. We
need to recompute chol($C_y$) to do this, but not chol($C_z$),
since $C_z$ depends only on $\theta_z$ but not $z$. We 
repeat this step for $m$ times (a tuning parameter) before moving
to the next hyperparameter in $\theta_z$.\vspace{-5pt}
\end{enumerate}
\end{enumerate}

In this scheme, the hyperparameters $\theta_y$ and $\theta_z$ are not
highly correlated and hence are relatively easy to sample using the
Metropolis algorithm. The latent variables $z$ are highly
correlated. Because updating the $z$-values is hard, we try to update
them as much as possible.  Notice that $C_z$ depends only on $x$ and
$\theta_z$, so a change of $z$ will not result in a change of
$C_z$. Hence once we update a component of $\theta_z$ (and obtain a new
$C_z$), it makes sense to do $m>1$ updates on $z$
before updating another $z$-hyperparameter.

\end{subsection}

\end{section}

\begin{section}{Experiments}\label{sec:exp}

We will compare our GPLC model with \citeauthor{Goldberg:1998}'s GPLV
model, and with a standard GP regression model having Gaussian
residuals of constant variance.

\begin{subsection}{Datasets}

We use four synthetic datasets, with one or three covariates, and
Gaussian or non-Gaussian residuals, as summarized below:

\begin{center}	
	\begin{tabular}{c|c|c|c|c}
		\hline
		Dataset & $p$ & True function & Residual SD & Residual distribution \\ \hline
		U1 & 1 & $f(x)$ & $r(x)$ &Gaussian \\\hline
		U2 & 1 & $f(x)$ & $r(x)$ &non-Gaussian \\\hline
		M1 & 3& $g(x)$ & $s(x)$ &Gaussian \\\hline
		M2 & 3& $g(x)$ & $s(x)$ &non-Gaussian\\\hline
	\end{tabular}\vspace{6pt}
\end{center}

Datasets U1 and U2 both have one covariate, which is uniformly drawn from
[0,1], and the true function is
	\[ f(x_i) = [1+\sin(4x_i)]^{1.1} \]

For U1, the response $y_i=f(x_i) + \epsilon_i$ is contaminated with a
Gaussian noise, $\epsilon_i$, with input-dependent standard deviation
	\[ SD(\epsilon_i) =r(x_i) = 0.2+0.3\exp[-30(x_i-0.5)^2] \]

For U2, the response has a non-Gaussian residual, $\omega$, with
a location-scale extreme value distribution, $EV(\mu, \sigma)$
\citep[see][]{GEV:2011}, with probability density
$(1/\sigma)e^{(\omega-\mu)/\sigma}\exp\left(-e^{(\omega-\mu)/\sigma}\right)$.
The mean of $\omega$ is $E(\omega) = \mu + \sigma \gamma$, 
where $\gamma=0.5772\ldots$ is the Euler's constant. The 
variance of $\omega$ is $\Var(\omega) = \pi^2\sigma^2/6$.
We translate and scale
the EV residuals so that their mean is zero and
their standard deviation is $r(x)$ (same as those of $\epsilon$ in U1).
The density curve of a EV residual with mean 0 and variance 1 is shown
in Figure \ref{fig:evdensity}.

\begin{figure}[b]
\centering
\includegraphics[height=2.4in,width=3.6in]{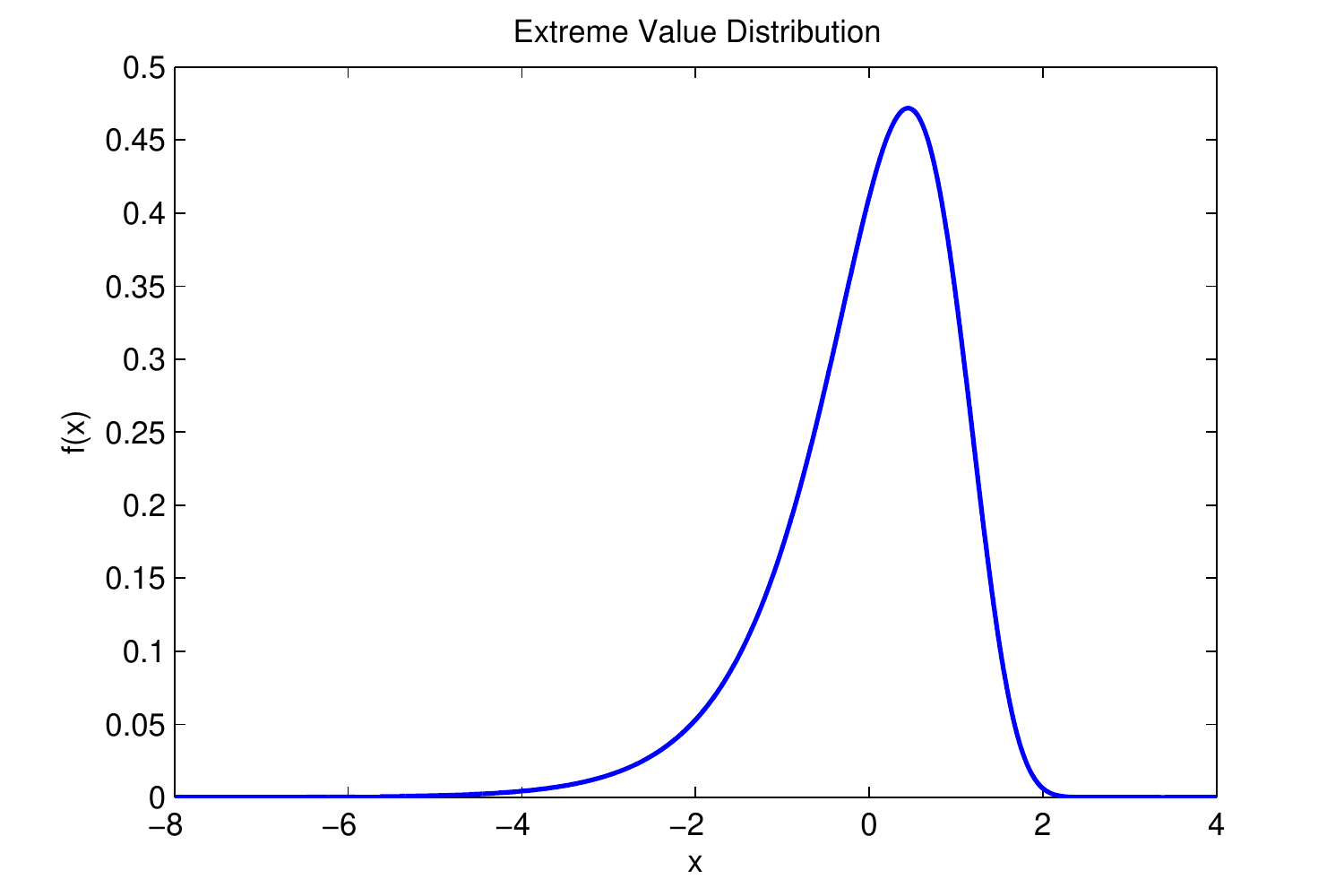}
\vspace{-5pt}
\caption{Density curve of extreme value residual with mean 0 and variance 1.}
\label{fig:evdensity}
\end{figure}

%

Datasets M1 and M2 both have three independent, standard normal
covariates, denoted as $x=(x_1,x_2,x_3)$.  The true function is
	\[ g(x) =[1+\sin(x_1/1.5+2)]^{0.9} - [1+\sin(x_2/2+x_3/3-2)]^{1.5} \]

As we did for U1 and U2, we add Gaussian residuals to M1 and extreme
value residuals to M2.  For both M1 and M2, the standard deviations of these 
residuals depend on the input covariates as follows:
		\[ s(x) =0.1 + 0.4\exp[-0.2(x_1-1)^2-0.3(x_2-2)^2] + 0.3\exp[-0.3(x_3+2)^2] \]

%
%

\end{subsection}

\begin{subsection}{Predictive performance of the methods}

For each dataset (U1, U2, M1, and M2), we randomly generated 10
different training sets (using the same program but different random
seeds), each with $n=100$ observations, and a test dataset with
$N=5000$ observations. We obtained MCMC samples using the methods
described in the previous section, dropping the initial 1/4 samples as
burn-in, and used them to make predictions for the test cases.


In order to evaluate how well each model does in terms of the mean
of its predictive distribution, we computed the mean
squared error (MSE) with respect to the true function values $f(x_i)$
as follows
	\begin{equation}
		\MSE(\hat{y}) = \frac{1}{N}\sum_{i=1}^{N} (\hat{y}_i-f(x_i))^2
		\label{eqn:mse}
	\end{equation}
where $\hat{y}_1,...,\hat{y}_N)$ are the predicted
responses for test cases. We also computed the average negative log-probability 
density (NLPD) of the responses in the test cases, as follows
	\begin{equation}
		\NLPD(\hat{y}) = -\frac{1}{N}\sum_{i=1}^{N} \log\left( \frac{1}{M}\sum_{j=1}^M\psi(y^{(i)}|\hat{\mu}_{ij},\hat{\sigma}^2_{ij})\right)
		\label{eqn:nlpd}
	\end{equation}
where $\psi(\cdot|\mu,\sigma^2)$ denotes the probability density for
$N(\mu,\sigma^2)$, $\hat{\mu}_{ij}, \hat{\sigma}^2_{ij}$ is the
predictive mean and variance for test case $y^{(i)}$ using the
hyperparameters and latent variables from the $j$th MCMC iteration, and 
$M$ is the number of MCMC samples used for prediction.

We give pairwise comparison of the MSE and the NLPD in Figures
\ref{fig:u1_resmat}, \ref{fig:u2_resmat}, \ref{fig:m1_resmat}, and
\ref{fig:m2_resmat}. These plots show that both GPLC and GPLV give
smaller NLPD values than the standard GP model for all datasets.  At
least for the multivariate datasets, GPLC and GPLV also give smaller
MSEs than the standard GP model. This shows that both GPLC and GPLV
can be effective for heteroscedastic regression problems.

Comparing GPLC and GPLV, we notice that for datasets with Gaussian
residuals, GPLC is almost as good as GPLV (except for the NLPDs for
U1, where GPLV gives smaller values 8 out of 10 times), while for
non-Gaussian residuals, GPLC is the clear winner, giving MSEs and
NLPDs that are smaller than for GPLV most of the time. The numerical
MSE and NLPD values are listed in the Appendix.

%

	\begin{figure}[p]

	\makebox[3.2in]{Negative Log Probability Density}
        \makebox[3.2in]{~~~~Mean Squared Error}

	\hspace*{-0.25in}\begin{minipage}{3.2in}
			\includegraphics[width=3.6in,height=3in]{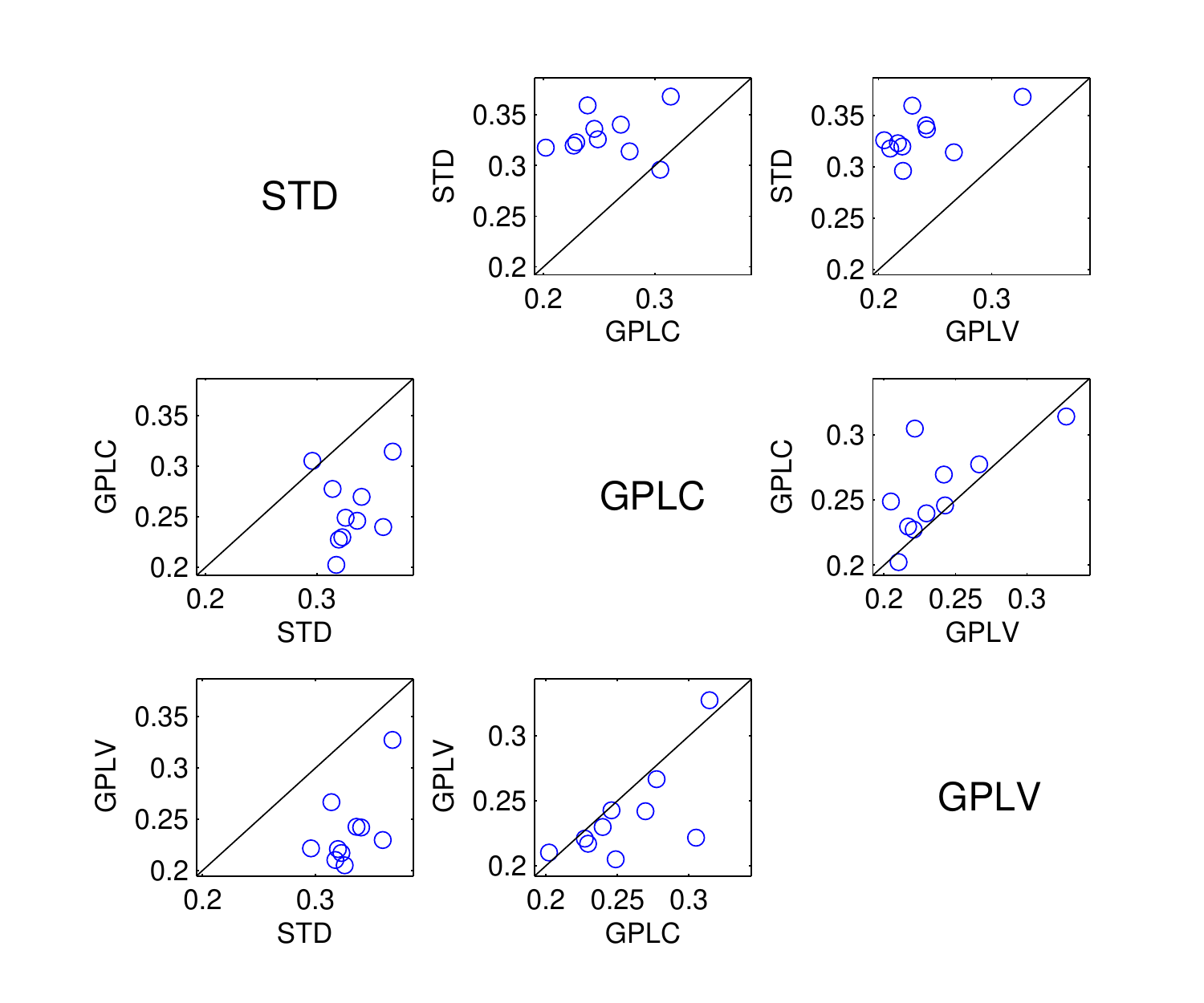}
	\end{minipage}
	\begin{minipage}{0.25in}
	\flushright
	\rule{.1pt}{3in}
	\end{minipage}
	\hspace{-0.1in}\begin{minipage}{3.2in}
			\includegraphics[width=3.6in,height=3in]{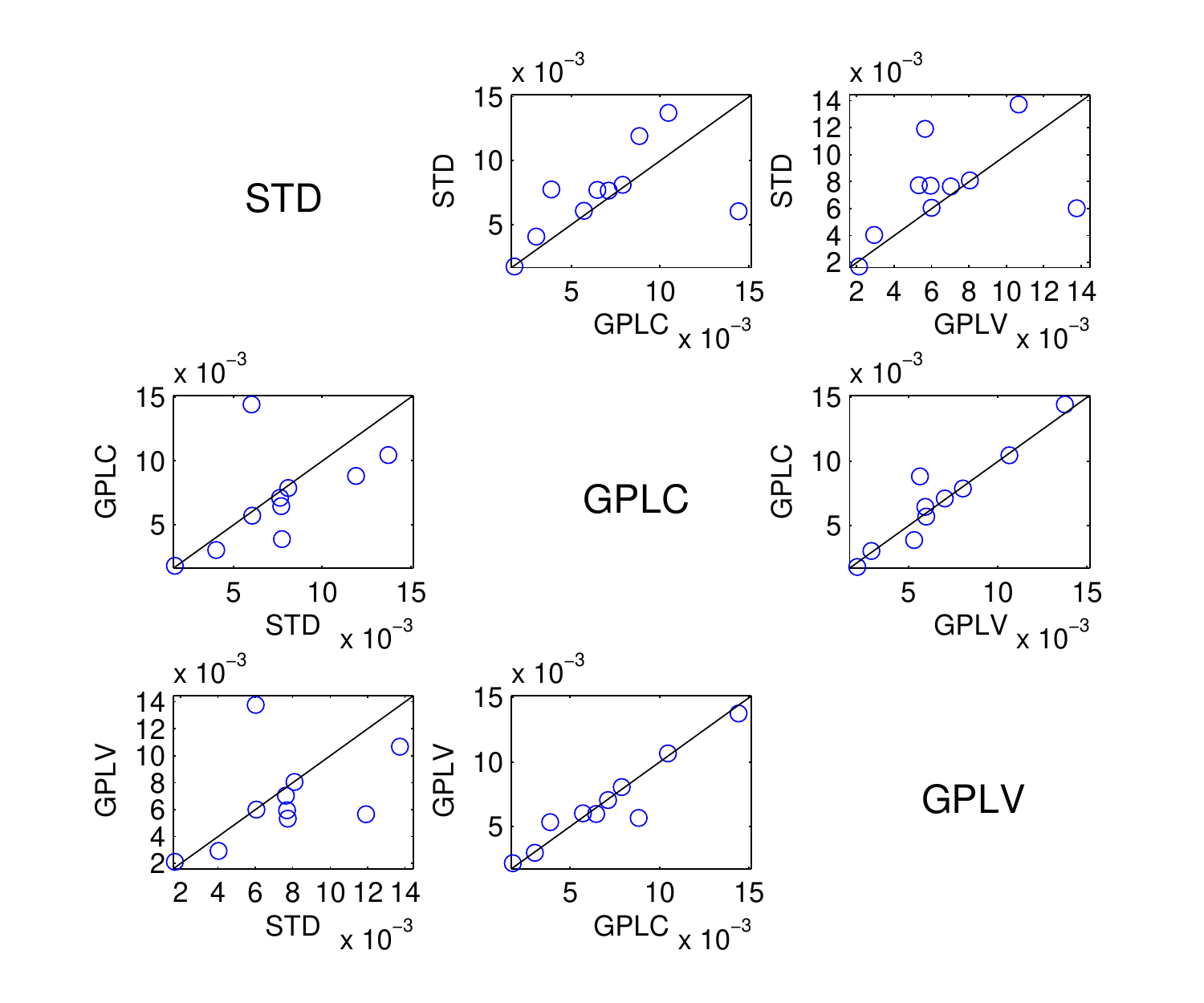}
	\end{minipage}
	\caption{Dataset U1:\ \ 
                 Pairwise comparison of methods using NLPD(Left) and MSE(right)}
	\label{fig:u1_resmat}
	\end{figure}

	\begin{figure}[p]

	\makebox[3.2in]{Negative Log Probability Density}
        \makebox[3.2in]{~~~~Mean Squared Error}

	\hspace*{-0.25in}\begin{minipage}{3.2in}	
			\includegraphics[width=3.6in,height=3in]{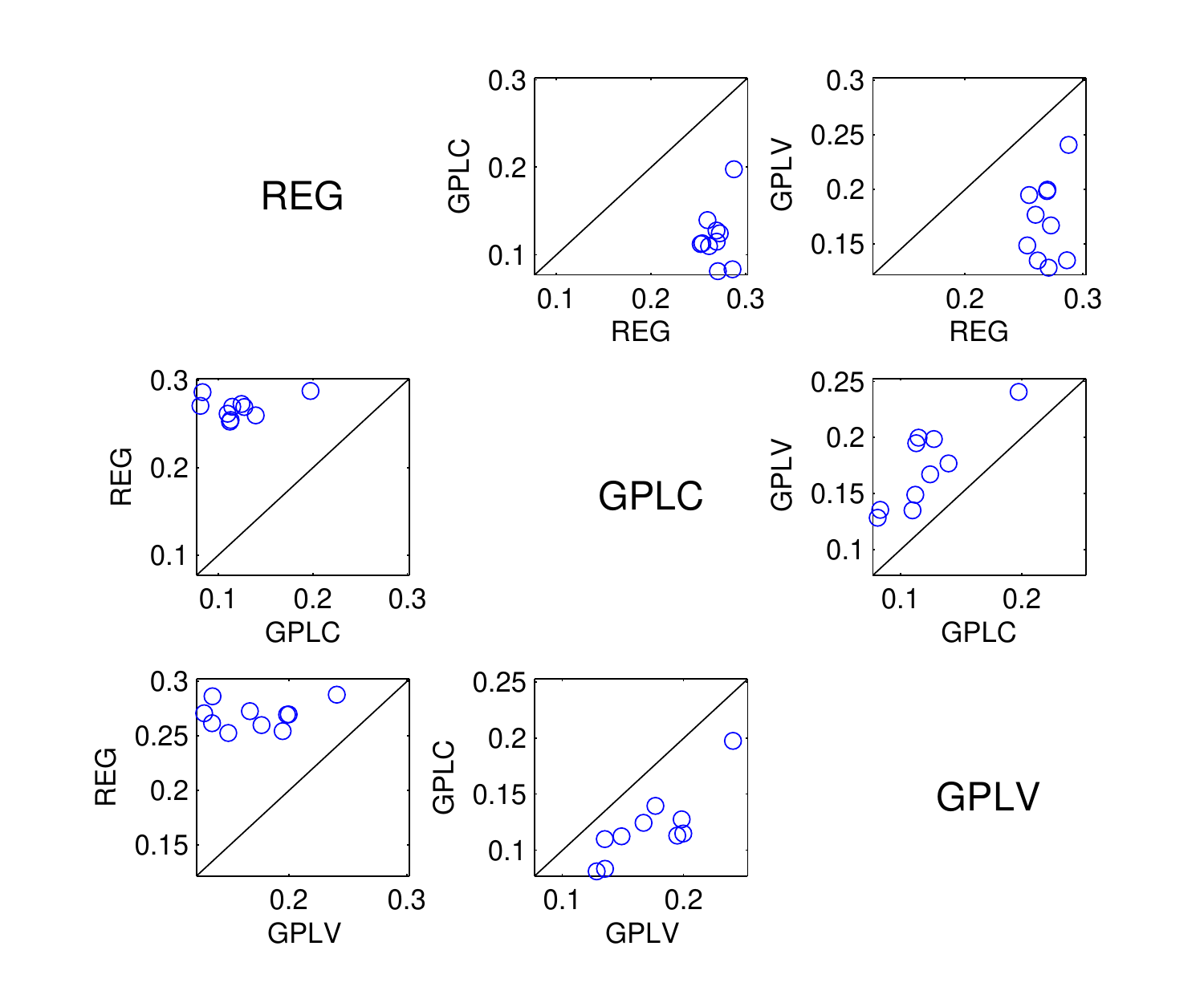}
	\end{minipage}
	\begin{minipage}{0.25in}
	\flushright
	\rule{.1pt}{3in}
	\end{minipage}
	\hspace{-0.1in}\begin{minipage}{3.2in}	
			\includegraphics[width=3.6in,height=3in]{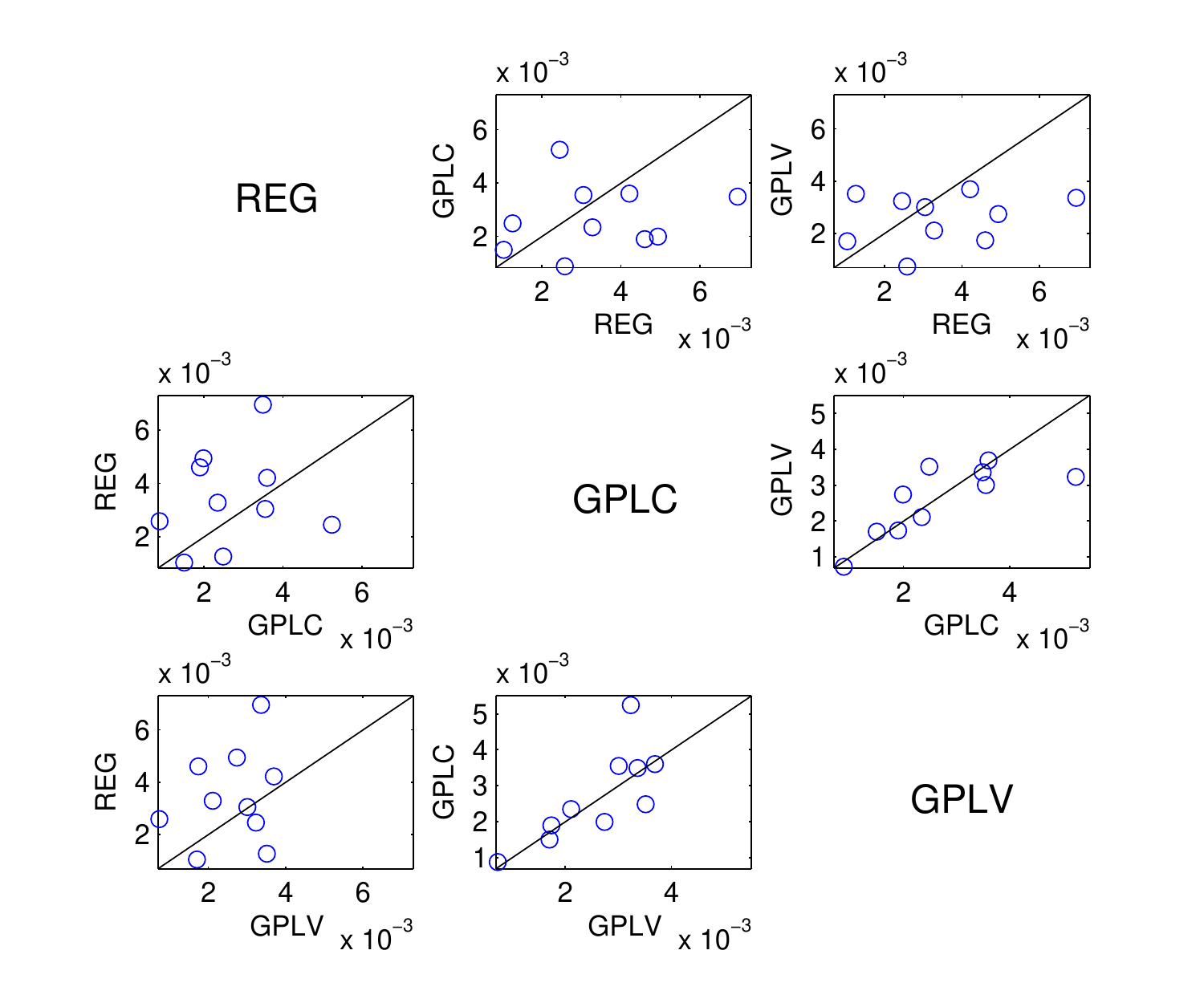}
	\end{minipage}
	\caption{Dataset U2:\ \ 
                 Pairwise comparison of methods using NLPD(Left) and MSE(right)}
	\label{fig:u2_resmat}

	\end{figure}

	\begin{figure}[p]

	\makebox[3.2in]{Negative Log Probability Density}
        \makebox[3.2in]{~~~~Mean Squared Error}

	\hspace*{-0.25in}\begin{minipage}{3.2in}	
			\includegraphics[width=3.6in,height=3in]{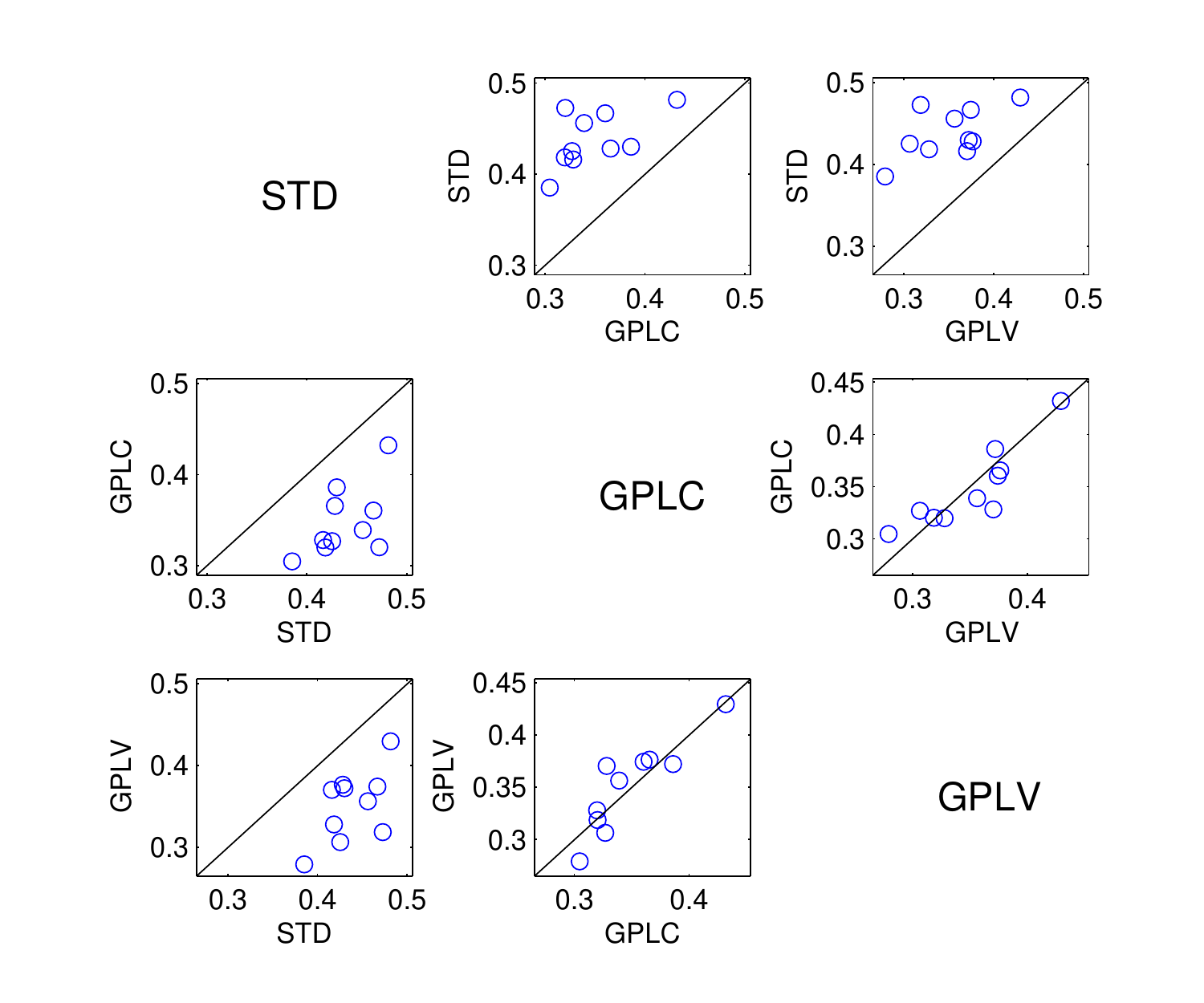}
	\end{minipage}
	\begin{minipage}{0.25in}
	\flushright
	\rule{.1pt}{3in}
	\end{minipage}
	\hspace{-0.1in}\begin{minipage}{3.2in}	
			\includegraphics[width=3.6in,height=3in]{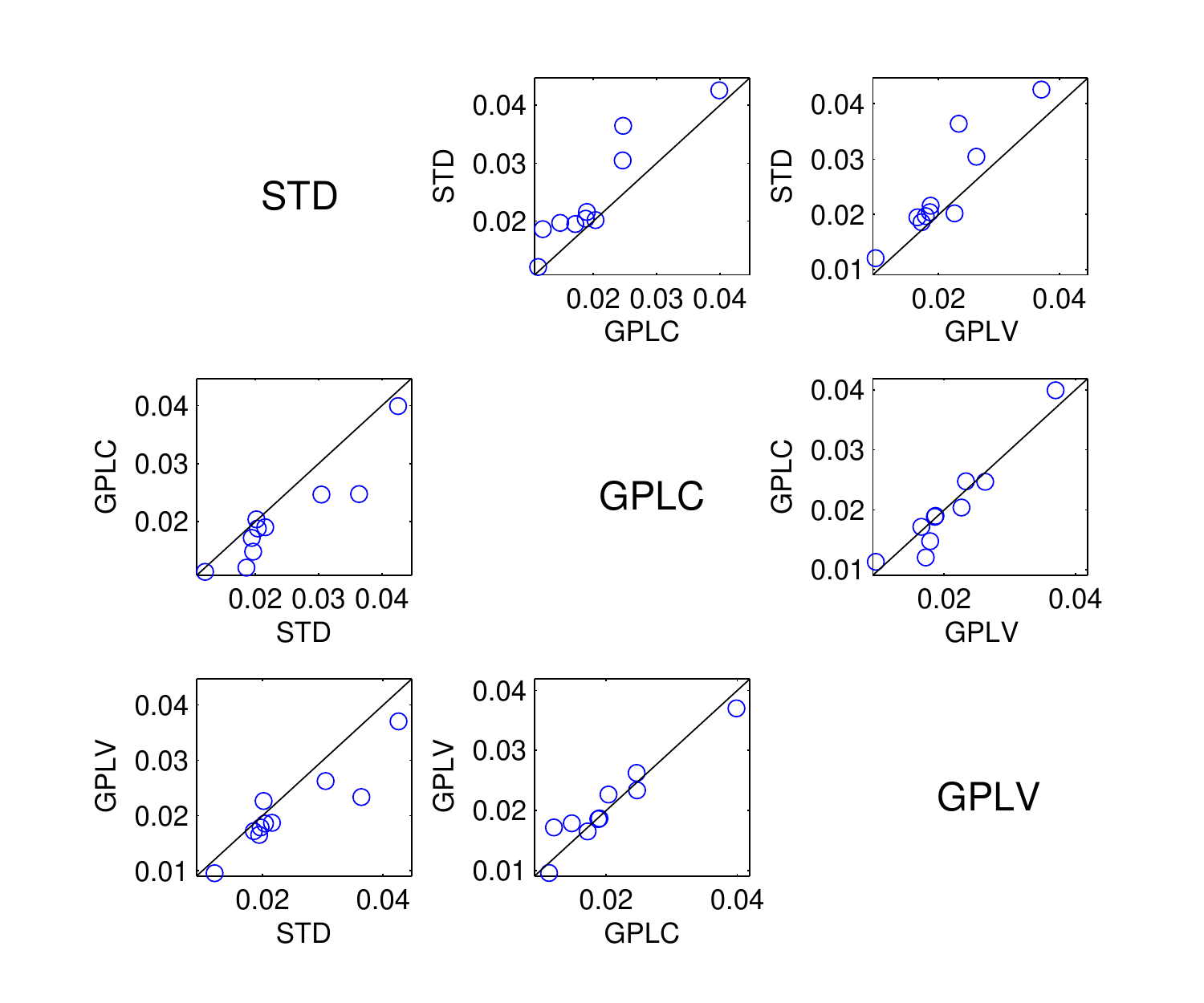}
	\end{minipage}
	\caption{Dataset M1:\ \ 
                 Pairwise comparison of methods using NLPD(Left) and MSE(right)}
	\label{fig:m1_resmat}
	\end{figure}

	\begin{figure}[p]

	\makebox[3.2in]{Negative Log Probability Density}
        \makebox[3.2in]{~~~~Mean Squared Error}

	\hspace*{-0.25in}\begin{minipage}{3.2in}	
			\includegraphics[width=3.6in,height=3in]{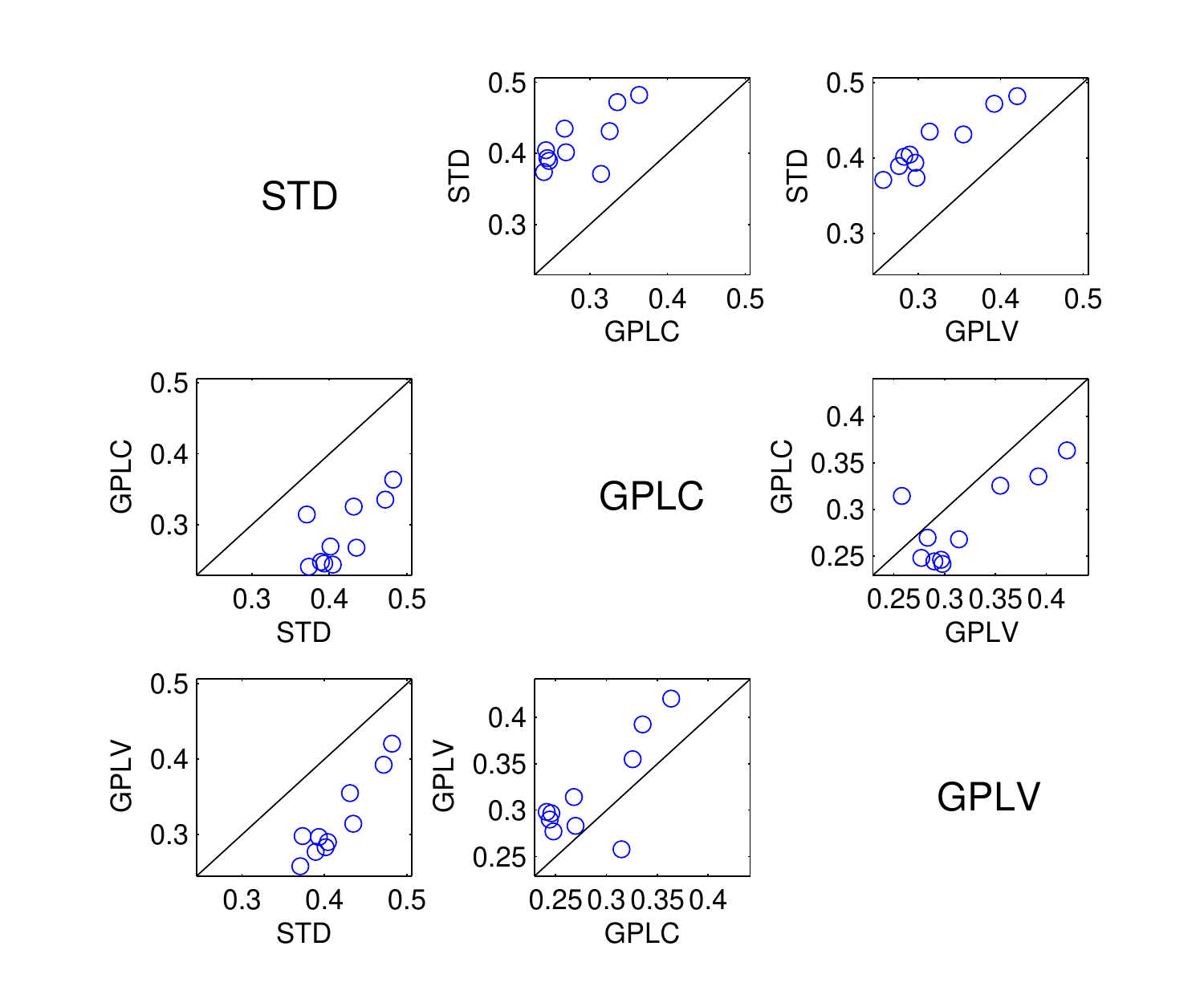}
	\end{minipage}
	\begin{minipage}{0.25in}
	\flushright
	\rule{.1pt}{3in}
	\end{minipage}
	\hspace{-0.1in}\begin{minipage}{3.2in}	
			\includegraphics[width=3.6in,height=3in]{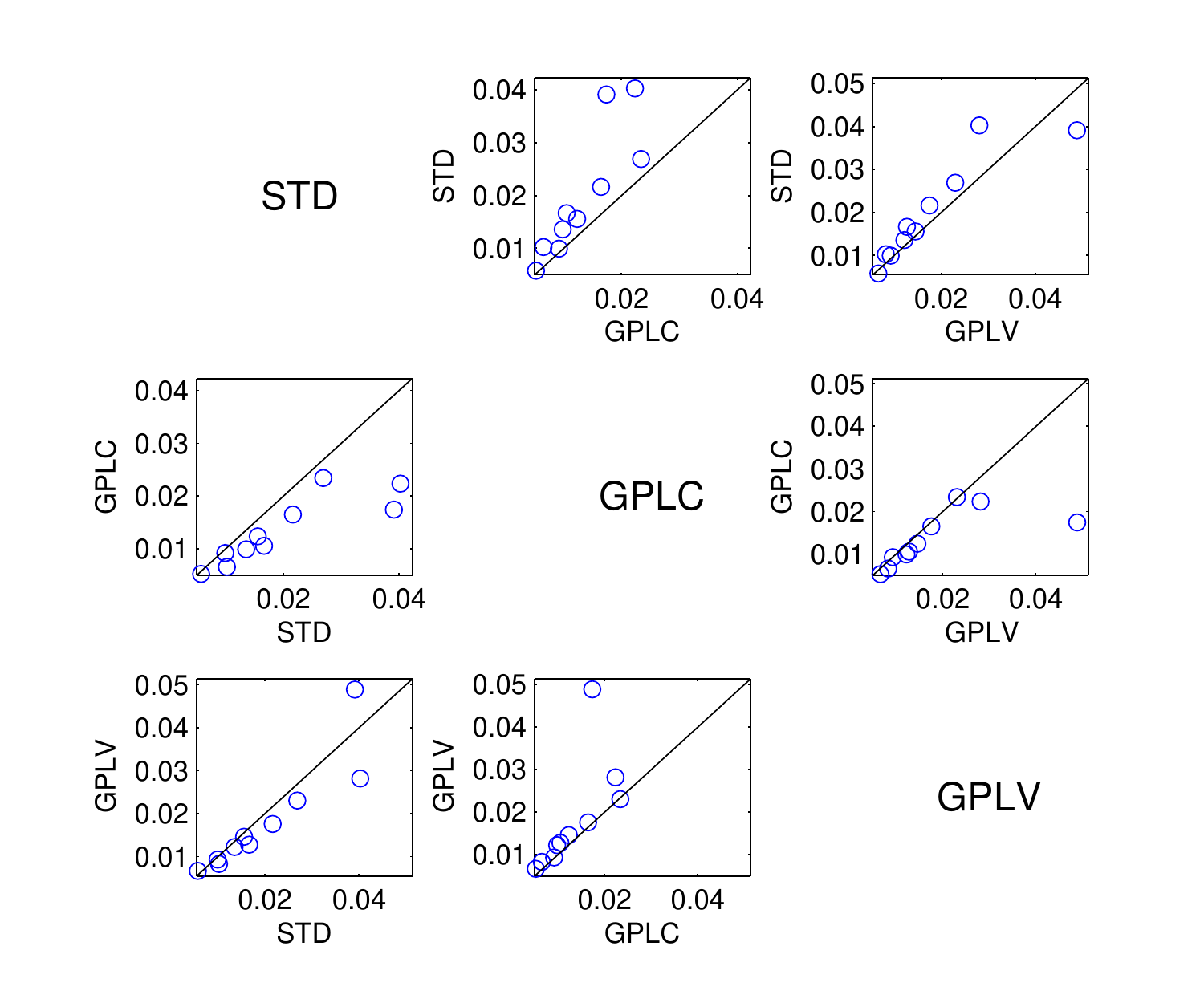}
	\end{minipage}
	\caption{Dataset M2:\ \ 
                 Pairwise comparison of methods using NLPD(Left) and MSE(right)}
	\label{fig:m2_resmat}
	\end{figure}

\end{subsection}

\begin{subsection}{Comparison of MCMC methods for GPLV models}
	
To test whether or not the modified Metropolis sampler described in
Section \ref{sec:modified_metro} is effective, we compare it to two
standard samplers, which update the latent values one by one using
either the Metropolis algorithm or the univariate step-out slice
sampler. The Metropolis algorithm is used to update the
hyperparameters in all of the three samplers. The significant
computations are listed in Table \ref{tbl:compacf_complexity}.

We adjust the tuning parameters so that the above samplers are
reasonably efficient.  For the slice sampler, we use a slice width of
1, and allow indefiite stepping-out. 
For the univariate Metropolis sampler, we adjust
the standard deviations of the Gaussian proposals so that the
acceptance rate of each parameter variable is around 50\%.  For the
modified Metropolis sampler, we set $a=0.3$ and $m=40$. The
acceptance rate for $z_i$ is around 1.4\%, but since we 
sample $z_i$ 40 times for each iteration, we have about 60\% of chance
to get a new value of $z_i$ while all hyperparameters are updated
once.


\begin{table}[b]
	\begin{tabular}{c|c|c|c|c}
	\hline
	\multicolumn{2}{c|}{} & $\theta_y$ & $\theta_z$ & $z$ \\ \hline		
	\multirow{2}{*}{Modified Metropolis} & Operation & $C_y$, chol$(C_y)$
													& $C_z$, chol$(C_z)$ 
													& $C_y$, chol$(C_y)$ \\ \cline{2-5}
	& \# of such operations & $p+1$ & $p+1$ & $m(p+1)$ \\\hline
	\multirow{2}{*}{Metropolis/Slice} & Operation & $C_y$, chol$(C_y)$
													& $C_z$, chol$(C_z)$ 
													& rank-1 up chol$(C_y)$ \\ \cline{2-5}
	& \# of such operations & $p+1$ & $p+1$ & $n$ \\\hline
	\end{tabular}
	\caption{Major operations of the MCMC methods for GPLV}
	\label{tbl:compacf_complexity}	
\end{table}

The efficiency of an MCMC method is usually measured by the
autocorrelation time, $\tau$, of values from the chain it 
produces \citep[see][]{Neal:1993}:
	\begin{equation}
		\tau = 1 + 2\sum_{i=1}^{\infty} \gamma_i
	\end{equation}
where $\gamma_i$ is the lag-$i$ autocorrelation. Roughly speaking, the
autocorrelation time is the number of iterations a sampler needs to
obtain another nearly uncorrelated sample.  

With an MCMC sample of size $M$, we can only estimate sample
autocorrelations $\hat{\gamma_i}$ up to $i=M-1$, and since these
estimates are all noisy, we usually estimate $\tau$ from a more
limited set of autocorrelations, as follows: \begin{equation}
\hat{\tau} = 1 + 2\sum_{i=1}^{k} \hat{\gamma_i} \end{equation} where
$k$ is a cut-off point where $\hat{\gamma_i}$ seems to be nearly 0 for
all $i>k$,

In our experiment, we will consider the autocorrelation time of both
the hyperparameters and the latent variables. The model has four
hyperparameters ($\eta_y,\rho_y$ and $\eta_z,\rho_z$), so it is not
too difficult to look at all of them. But there are $n=100$ latent
variables, each with its own autocorrelation time. Instead of
comparing all of them one by one, we will compare the autocorrelation
time of the sum of the latent variables as well as the sum of the
squares of the latent variables.


Another measure of sampler efficiency is the time it takes for a
sampler to reach equilibrium, i.e. the stationary distribution of the
Markov chain. This is often referred to as the Markov chain ``mixing
time'' (denoted as $T_M$). Theoretical bound of mixing rate can be
found for some classical algorithms, however, in practice, the mixing
time of a sophisticated sampler is usually very difficult to
determine. It is common practice to look at trace plots of
log-probability density values to decide whether or not a chain has
reached equilibrium (this is usually how the number of ``burn-in''
iterations is decided).

The mixing time and the autocorrelation time usually agree in the
sense that a sampler that takes less time to get a new uncorrelated
sample can usually achieve equilibrium faster, and vice versa, though
this is not always the case. 

Note both autocorrelation time $\tau$ and mixing time $T_M$ take the
number of iterations of the a Markov chain as their unit of
measurement. However, the CPU time required for an iteration differs
between samplers. For a fair comparison, we adjust $\tau$ by
multiplying it with the average CPU time per iteration. The result,
which we denote as $\tilde{\tau}$, measures the CPU time a sampler
needs to obtain an uncorrelated sample.  To fairly compare mixing
times using trace plots, we will adjust the number of iterations in
the plots so that each entire trace takes the same amount of time.

We run the three samplers five times, starting from the same point
(the prior mean) but with different random seeds. The average adjusted
autocorrelation times are listed in Table \ref{tbl:compacf}. The
modified Metropolis sampler significantly outperforms the others at
sampling the latent variables: it is about 50 to 100 times faster than
the regular Metropolis sampler and slice sampler. For the
hyperparameters, however, the modified Metropolis sampler gives
roughly the same autocorrelation times as the regular Metropolis
sampler does. Both of them seems to work better than slice sampler,
but the difference is much smaller than the difference in sampling
latent variables.  Figure \ref{fig:compacf_acf} shows selected
autocorrelation plots from one of the five runs (adjusted for
computation time).

	\begin{table}[t]
	\centering
	\begin{tabular}{r|r|r|r|r|r|r}
	\hline
	 & \multicolumn{6}{|c}{$\tilde{\tau}$} \\ \cline{2-7}
	 & $\eta_y$~~ & $\rho_y$~~ & $\eta_z$~~ & $\rho_z$~~ & $\sum_i z_i$ &$\sum_i z_i^2$ \\ \hline
	Modified Metropolis	&	3.06	&	4.92	&	3.09	&	10.63	&	0.32	&	0.46	\\ \hline
Metropolis	&	2.81	&	11.21	&	2.73	&	27.26	&	13.78	&	13.92	\\ \hline
Slice	&	13.37	&	16.71	&	11.85	&	23.65	&	37.81	&	53.81	\\ \hline

	\end{tabular}
	\caption{Autocorrelation times (adjusted for computation time) 
	of MCMC methods for GPLV.}
	\label{tbl:compacf}
	\end{table}

Figure \ref{fig:compacf_trace} gives the trace plots of the three
methods for one of the five runs (other runs are similar).
The bottom plot is the trace of log-probability density
values of the initial 400 iterations of the slice sampler. The middle
plot shows the initial iterations of the regular Metropolis sampler,
with the number adjusted to take the same time as the 400 slice
sampler iterations. The top plot shows the initial iterations of
the modified Metropolis sampler, again taking the same computation time. 

It is clear that the modified Metropolis takes the least time to
mix. Starting from the prior mean (which seem to be a reasonable
initial point), with log-probability density (LPD) value of
approximately $-13$, the modified Metropolis method immediately pushes
the LPD to 70 at the second step, and then soon declines slightly to
what appears to be the equilibrium distribution. The other two methods
move take much more time to reach this equilibrium distribution, with
the simple Metropolis and slice sampling methods taking roughly the
same amount of time. 

We conclude that the modified Metropolis is the best of these MCMC
method --- the fastest to reach equilibrium, the best at sampling
latent values thereafter, and at least as good at sampling
hyperparameters.

	\begin{figure}[p]
		\centering
		\includegraphics[height=4in,width=7in]{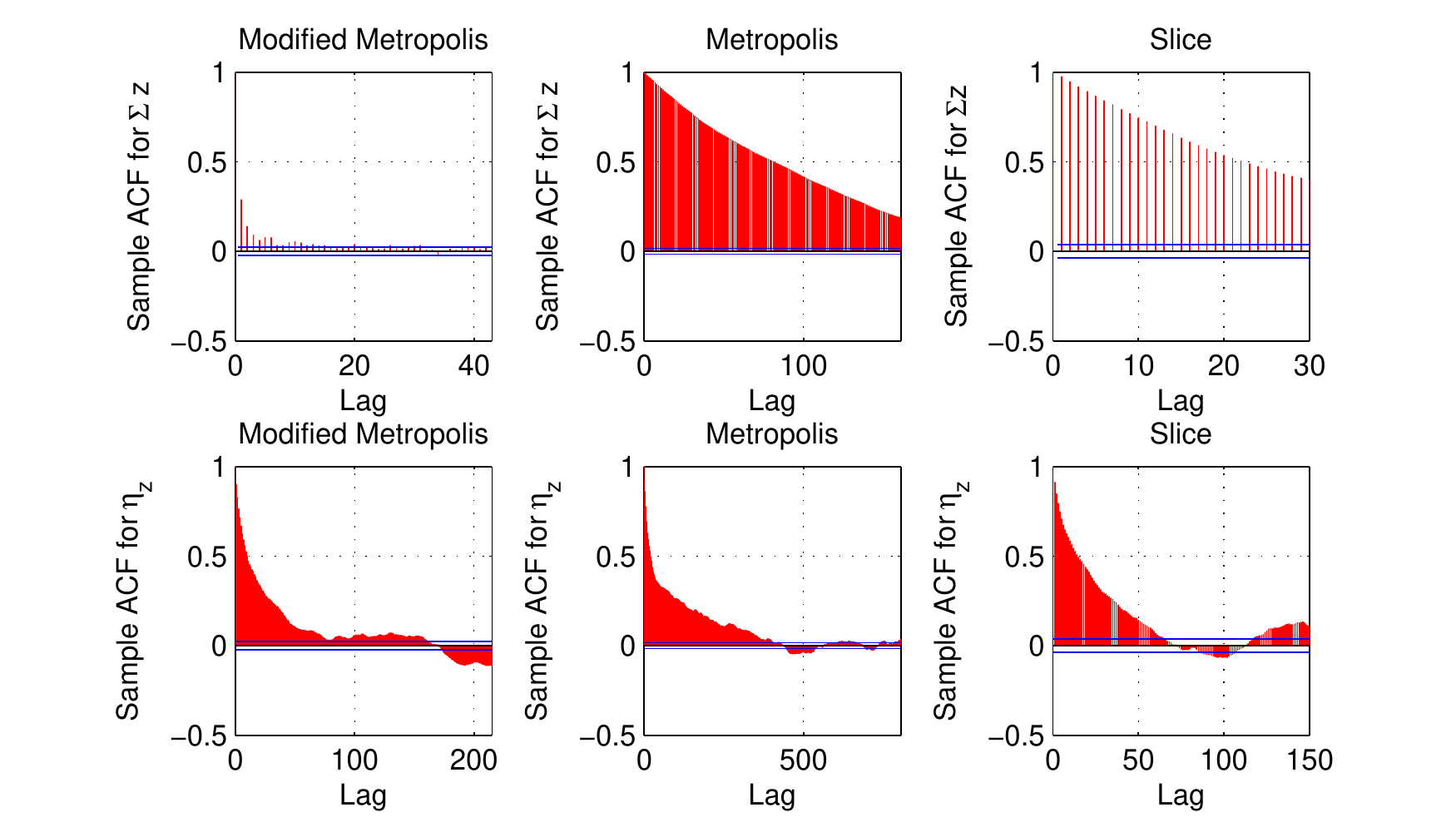}
                \vspace{-0.3in}

		\caption{Selected autocorrelation plots for MCMC methods for 
                         GPLV (with horizontal scales that adjust for 
                         computation time).}
		\label{fig:compacf_acf}
	\end{figure}
	\begin{figure}[p]
		\centering
		\includegraphics[height=4in,width=7in]{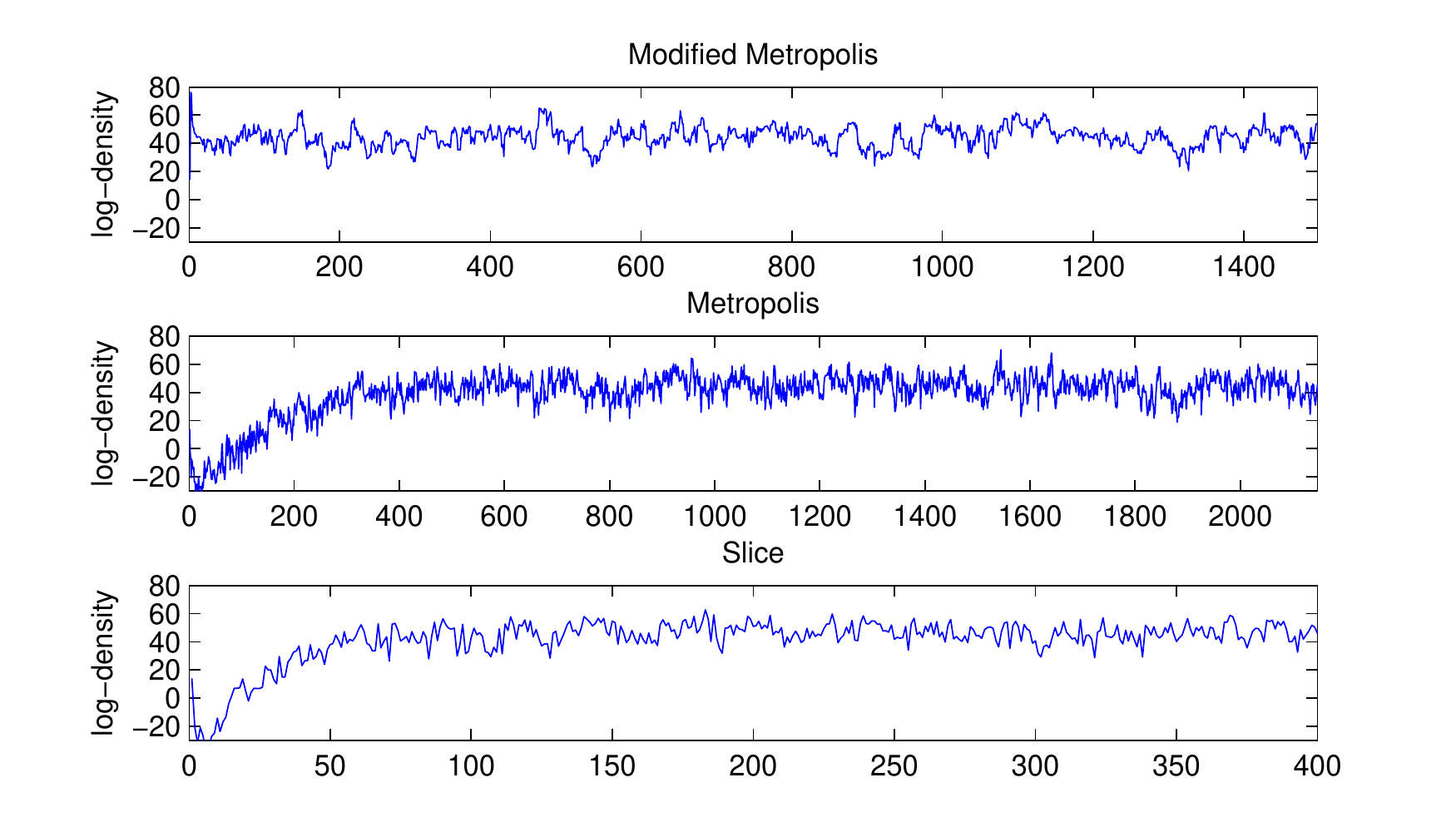}
                \vspace{-0.3in}

		\caption{Trace plots of log posterior density for MCMC 
                         methods for GPLV.}
		\label{fig:compacf_trace}
	\end{figure}

In a simpler context --- financial time series where no ``main'' GP is
needed, since the mean response can be taken to be always zero ---
\cite{Wilson:2010} use the ``elliptical slice sampling'' method of
\cite{ell} to sample latent values that determine the variances of
observations.  Elliptical slice sampling is related to the modified
Metropolis method above.  It would be interesting to see how they
compare in a general regression context.

\end{subsection}

\end{section}
%
%
%

\newpage
\section*{Appendix:\ \ MSE and NLPD for each method and training set}

\fontfamily{pcr}

\noindent
\large \makebox[1.1in][l]{Dataset U1:} \small \hspace{0.1in} 
\begin{tabular}{c||c|c|c||c|c|c}
		\hline
			 & \multicolumn{3}{|c||}{NLPD} & \multicolumn{3}{c}{MSE} \\ \cline{2-7}
\!\!\!\!Training Set\!\!&REG&GPLC&GPLV&REG&GPLC&GPLV \\ \hline
\multicolumn{7}{c}{} \\[0.1in] \hline
\makebox[0.9in]{1}
	&	0.3364	&	\textbf{0.2459}	&	0.2480	&	0.0119	&	0.0088	&	\textbf{0.0059}	\\ \hline
2	&	0.3259	&	0.2489	&	\textbf{0.2076}	&	0.0077	&	\textbf{0.0039}	&	0.0058	\\ \hline
3	&	0.3142	&	0.2774	&	\textbf{0.2661}	&	\textbf{0.0060}	&	0.0144	&	0.0139	\\ \hline
4	&	0.3198	&	0.2273	&	\textbf{0.2237}	&	0.0077	&	0.0065	&	\textbf{0.0058}	\\ \hline
5	&	0.3231	&	0.2296	&	\textbf{0.2185}	&	0.0076	&	\textbf{0.0071}	&	0.0073	\\ \hline
6	&	0.3596	&	0.2397	&	\textbf{0.2321}	&	0.0137	&	\textbf{0.0105}	&	0.0110	\\ \hline
7	&	0.3404	&	0.2696	&	\textbf{0.2374}	&	0.0040	&	\textbf{0.0030}&	\textbf{0.0030}	\\ \hline
8	&	0.3683	&	\textbf{0.3143}	&	0.3305	&	0.0081	&	\textbf{0.0079}	&	0.0080	\\ \hline
9	&	0.3177	&	\textbf{0.2023}	&	0.2172	&	0.0061	&	0.0057	&	\textbf{0.0055}	\\ \hline
10	&	0.2961	&	0.3050	&	\textbf{0.2107}	&	\textbf{0.0017}	&	0.0018	&	0.0020	\\ \hline

		\end{tabular}

\vspace{0.2in}

\noindent
\large \makebox[1.1in][l]{Dataset U2:} \small \hspace{0.1in} 
\begin{tabular}{c||c|c|c||c|c|c}
		\hline
\makebox[0.9in]{1}
	&	0.2878	&	\textbf{0.1976}	&	0.2408	&	0.0070	&	0.0035	&	\textbf{0.0034}	\\ \hline
2	&	0.2616	&	\textbf{0.1099}	&	0.1349	&	0.0026	&	0.0009	&	\textbf{0.0007}	\\ \hline
3	&	0.2527	&	\textbf{0.1123}	&	0.1488	&	\textbf{0.0013}	&	0.0025	&	0.0035	\\ \hline
4	&	0.2697	&	\textbf{0.1148}	&	0.1998	&	0.0042	&	\textbf{0.0036}	&	0.0037	\\ \hline
5	&	0.2695	&	\textbf{0.1275}	&	0.1985	&	\textbf{0.0030}	&	0.0036	&	\textbf{0.0030}	\\ \hline
6	&	0.2599	&	\textbf{0.1396}	&	0.1769	&	\textbf{0.0025}	&	0.0052	&	0.0032	\\ \hline
7	&	0.2544	&	\textbf{0.1130}	&	0.1948	&	\textbf{0.0010}	&	0.0015	&	0.0017	\\ \hline
8	&	0.2708	&	\textbf{0.0811}	&	0.1283	&	0.0046	&	0.0019	&	\textbf{0.0017}	\\ \hline
9	&	0.2863	&	\textbf{0.0833}	&	0.1353	&	0.0049	&	\textbf{0.0020}	&	0.0027	\\ \hline
10	&	0.2727	&	\textbf{0.1245}	&	0.1670	&	0.0033	&	0.0023	&	\textbf{0.0021}	\\ \hline

		\end{tabular}

\vspace{0.2in}

\noindent
\large \makebox[1.1in][l]{Dataset M1:} \small \hspace{0.1in} 
\begin{tabular}{c||c|c|c||c|c|c}
		\hline
\makebox[0.9in]{1}
	&	0.4726	&	\textbf{0.3202}	&	0.3407	&	0.0186	&	\textbf{0.0120}	&	0.0201	\\\hline
2	&	0.3852	&	0.3046	&	\textbf{0.2860}	&	0.0121	&	0.0113	&	\textbf{0.0098}	\\\hline
3	&	0.4560	&	\textbf{0.3390}	&	0.3410	&	0.0305	&	\textbf{0.0247}	&	0.0277	\\\hline
4	&	0.4300	&	0.3860	&	\textbf{0.3751}	&	0.0204	&	\textbf{0.0188}	&	0.0201	\\\hline
5	&	0.4817	&	0.4321	&	\textbf{0.3906}	&	0.0426	&	0.0399	&	\textbf{0.0351}	\\\hline
6	&	0.4668	&	0.3603	&	\textbf{0.3024}	&	0.0364	&	0.0247	&	\textbf{0.0163}	\\\hline
7	&	0.4282	&	\textbf{0.3656}	&	0.3799	&	0.0195	&	0.0172	&	\textbf{0.0171}	\\\hline
8	&	0.4184	&	\textbf{0.3198}	&	0.4022	&	0.0197	&	\textbf{0.0148}	&	0.0217	\\\hline
9	&	0.4161	&	\textbf{0.3281}	&	0.3817	&	\textbf{0.0202}	&	0.0204	&	0.0297	\\\hline
10	&	0.4253	&	0.3269	&	\textbf{0.3201}	&	0.0216	&	\textbf{0.0190}	&	0.0197	\\\hline

		\end{tabular}

\vspace{0.2in}

\noindent
\large  \makebox[1.1in][l]{Dataset M2:} \small \hspace{0.1in} 
\begin{tabular}{c||c|c|c||c|c|c}
		\hline
\makebox[0.9in]{1}
	&	0.3736	&	\textbf{0.2413}	&	0.298	&	0.0099	&	\textbf{0.0092}	&	0.0093	\\ \hline
2	&	0.3934	&	\textbf{0.2458}	&	0.2965	&	0.0136	&	\textbf{0.0099}	&	0.0122	\\ \hline
3	&	0.4015	&	\textbf{0.2695}	&	0.2832	&	0.0167	&	\textbf{0.0105}	&	0.0128	\\ \hline
4	&	0.4819	&	\textbf{0.3636}	&	0.4202	&	0.0391	&	\textbf{0.0174}	&	0.0489	\\ \hline
5	&	0.4311	&	\textbf{0.3257}	&	0.3548	&	0.0269	&	0.0234	&	\textbf{0.0230	}\\ \hline
6	&	0.4348	&	\textbf{0.2678}	&	0.3140	&	0.0216	&	\textbf{0.0165}	&	0.0176	\\ \hline
7	&	0.3892	&	\textbf{0.2479}	&	0.2770	&	0.0102	&	\textbf{0.0065}	&	0.0083	\\ \hline
8	&	0.3709	&	0.3147	&	\textbf{0.2580}	&	0.0058	&	\textbf{0.0052}	&	0.0067	\\ \hline
9	&	0.4043	&	\textbf{0.2441}	&	0.2899	&	0.0156	&	\textbf{0.0124}	&	0.0146	\\ \hline
10	&	0.4718	&	\textbf{0.3355}	&	0.3923	&	0.0403	&	\textbf{0.0223}	&	0.0281	\\ \hline

		\end{tabular}

\end{document}